\documentclass[letterpaper]{sig-alternate}
\usepackage{amsmath}
\usepackage{dsfont}

\newcommand{\coll}[1]{\textsf{#1}}
\newcommand{\set}[1]{\textsf{#1}}
\newcommand{\algname}[1]{\small{\textsc{#1}}\normalsize}

\newcommand{\term}[1]{\texttt{#1}}
\long\def\comment#1{}
\newcommand{\secref}[1]{Section~\ref{#1}}
\newcommand{\figref}[1]{Figure~\ref{#1}}
\newcommand{\tabref}[1]{Table~\ref{#1}}
\newcommand{\eqrefx}[1]{Eq.~\ref{#1}}
\newcommand{\remove}[1]{}

\begin{document}
\conferenceinfo{KDD'10,} {July 25--28, 2010, Washington, DC, USA.} 

\CopyrightYear{2010}

\crdata{978-1-4503-0055-1/10/07}

\clubpenalty=10000

\widowpenalty = 10000 

\title{Growing a Tree in the Forest: Constructing Folksonomies by Integrating Structured Metadata}

\numberofauthors{3}

\author{
%
\alignauthor Anon Plangprasopchok\\
       \affaddr{USC Information Sciences Institute}\\
       \affaddr{Marina del Rey, CA, USA}\\
       \email{plangpra@isi.edu}
\alignauthor Kristina Lerman\\
       \affaddr{USC Information Sciences Institute}\\
       \affaddr{Marina del Rey, CA, USA}\\
       \email{lerman@isi.edu}
\alignauthor Lise Getoor\\
       \affaddr{Department of Computer Science}\\ 
       \affaddr{University of Maryland}\\
       \affaddr{College Park, MD, USA}\\
       \email{getoor@cs.umd.edu}       
}

\maketitle
\begin{abstract}
Many social Web sites allow users to annotate the content with descriptive metadata, such as tags, and more recently to organize content hierarchically.  These types of structured metadata provide valuable evidence for learning how a community organizes knowledge. For instance, we can aggregate many personal hierarchies into a common taxonomy, also known as a folksonomy, that will aid users in visualizing and browsing social content, and also to help them in organizing their own content. However, learning from social metadata presents several challenges, since it is sparse, shallow, ambiguous, noisy, and inconsistent.  We describe an approach to folksonomy learning based on relational clustering, which \comment{that addresses these challenges by} exploits structured metadata contained in personal hierarchies. Our approach clusters similar hierarchies using their structure and tag statistics, then  incrementally weaves them into a deeper, bushier tree. We study folksonomy learning using social metadata extracted from the photo-sharing site Flickr, and demonstrate that the proposed approach addresses the challenges. Moreover, comparing to previous work, the approach produces larger, more accurate folksonomies, and in addition, scales better.
\comment{
We evaluate the learned folksonomy qualitatively and quantitatively by automatically comparing it to a reference taxonomy created by the Open Directory Project, and by manual labeling.  Our empirical results suggest that the proposed framework, which addresses the challenges listed above, improves significantly on existing folksonomy learning methods.		
}
\end{abstract}


\section{Introduction}
The social Web has changed the way people create and use information. Sites like Flickr, Del.icio.us, YouTube, and others, allow users to publish and organize content by annotating it with descriptive keywords, or tags. Some web sites also enable users to organize content hierarchically. The photo-sharing site Flickr, for example, allows users to group related photos in sets, and related sets in collections. Although these types of social metadata lack formal structure, they capture the collective knowledge of Social Web users. Once mined from the traces left by many users, such collective knowledge will add a rich semantic layer to the content of the Social Web that will potentially support many tasks in information discovery such as personalization, data mining, and information management.

A community's knowledge can be expressed through a common taxonomy, also called a \emph{folksonomy}, that is learned from social metadata created by many users. Compared to existing \remove{classification} hierarchies, such as Linnaean  classification system or WordNet, automatically learned folksonomies are attractive because they (1) represent collective agreement of many individuals; (2) are relatively inexpensive to obtain; (3) can adapt to evolving vocabularies and community's information needs; and (4) they are \remove{grounded in annotated content.} directly tied to the annotated content. A folksonomy can facilitate browsing of user-generated content, and help users visualize how their own content fits within the community's or aid them in organizing it.

Learning a folksonomy by integrating structured metadata created by many users presents a number of challenges. Since users are free to annotate data according to their own preferences, social metadata is  \emph{noisy}, \emph{shallow}, \emph{sparse}, \emph{ambiguous}, \emph{conflicting}, \emph{multi-faceted}, and expressed at \emph{inconsistent granularity levels} across many users.
Several recent works have addressed some of the above challenges. For instance, \cite{Heymann06, SchmitzTagging06} proposed inducing folksonomies from tags by utilizing tag statistics. The basic motivation behind these approaches is that more frequent tags describe more general concepts. However, frequency-based methods cannot distinguish between more general and more popular concepts. In our previous work, \algname{sig}~\cite{www09folksonomies}, we overcame this problem by using user-specified relations, extracted from personal hierarchies. Nevertheless, it ignored other evidence, e.g., structure of hierarchies and tags, which potentially address the challenges listed above.

We propose a novel approach to learn folksonomies \remove{by exploiting} from social metadata in the form of tags and user-specified shallow hierarchies. Our approach is driven by a similarity measure that utilizes statistics of both kinds of metadata to incrementally weave individual hierarchies into a deeper, more complete folksonomy. 
The approach has several advantages over previous work. Specifically, it: (1) better addresses the challenges of sparse, shallow, ambiguous, noisy and inconsistent data; (2) the approach is more scalable, especially when the learned folksonomies are deep; (3) it produces more consistent and richer folksonomies. We demonstrate the utility of our present approach on real-world data from Flickr, and introduce a simple metric, which evaluates the quality of the learned folksonomies in terms of depth and bushiness.
\comment{The approach demonstrates several advantages over the previous work in that, it addresses the social metadata challenges better and more scalable especially when the depth of the output tree is high. Moreover, it can produce more consistent and much richer folksonomies, according to the case study on the data set obtained from Flickr. We also introduce   
a simple metric, which takes both tree's bushiness and depth into consideration, to evaluate a tree structure. 
}

\comment{
The paper is organized as follows. In Section~\ref{sec:metadata} we describe the structured social metadata we use as evidence for learning folksonomies. \comment{Although we illustrate our approach and apply it to data extracted from the social photo-sharing site Flickr, it is general, and applies to other sites where users create personal hierarchies to organize their content or metadata.} In Section~\ref{sec:challenges} we list and illustrate the challenges in constructing folksonomies from metadata specified by many individual users. We present our method in Section~\ref{sec:approach} and evaluate it in Section~\ref{sec:results}.
}

\section{Structured Social Metadata}
\label{sec:metadata}
\comment{
Tags are keywords freely chosen by users from uncontrolled personal vocabularies to describe content.
} 
In addition to tagging content, some social Web sites also allow users to organize it hierarchically. \emph{Delicious} users can group related tags into bundles, and \emph{Flickr} users can group related photos into \emph{sets} and then group related sets in \emph{collections}. While the sites themselves do not impose any constraints on the vocabulary or semantics of the hierarchies, in practice users employ them to represent both subclass relationships (`dog' is a kind of `mammal') and part-of relationship (`my kids' is a part of `family'). Users appear to express both types of  relations (and possibly others) through personal hierarchies, in effect using the hierarchies to specify broader/narrower relations.
Even without strict semantics being attached to these relations, we believe that \remove{user-specified relations} personal hierarchies represent a novel, rich source of evidence for learning folksonomies\remove{ that is superior to using tags alone}.

\begin{figure}[tbh]
\begin{tabular}{cccc}
\multicolumn{2}{c}{\includegraphics[width=3.2in]{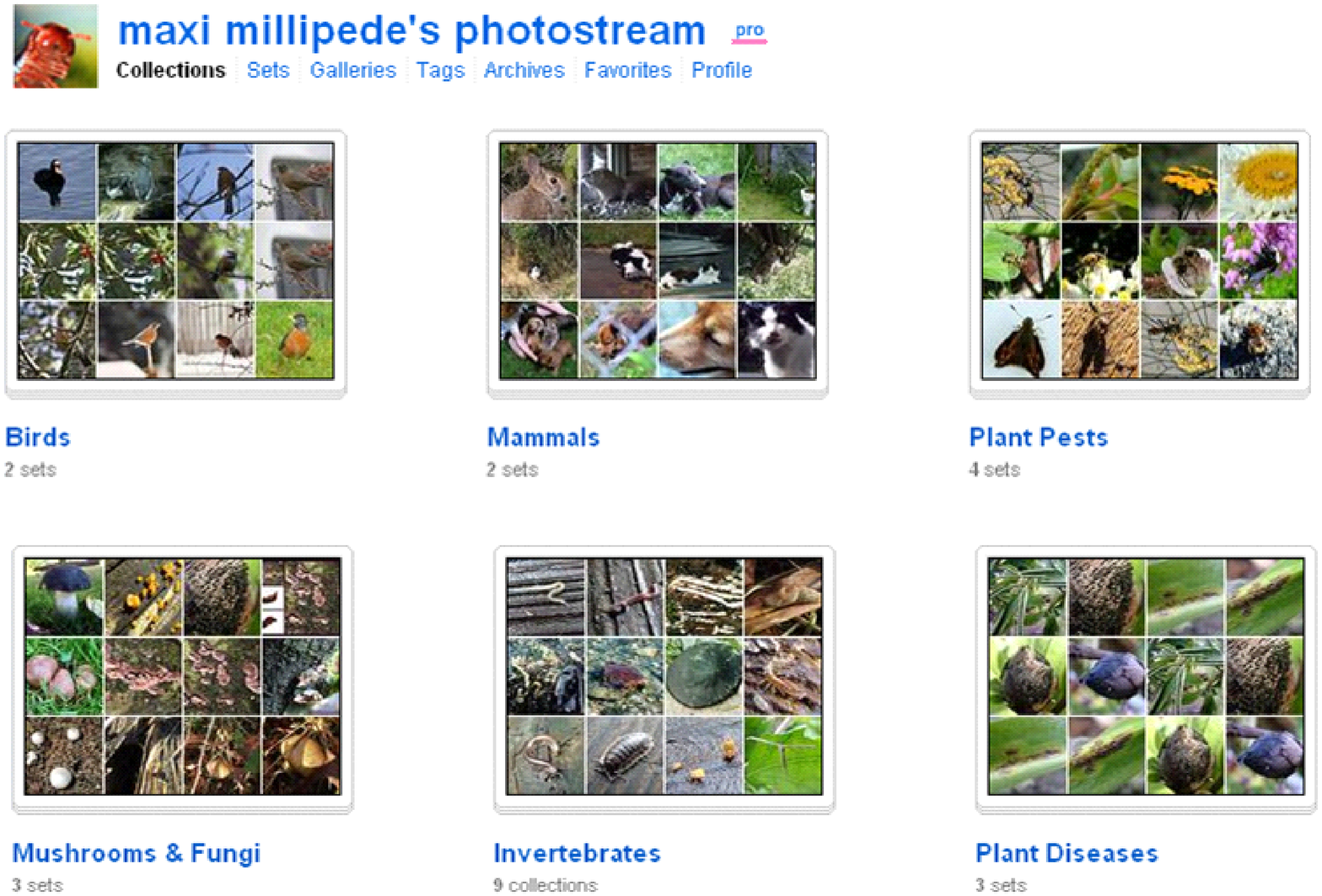}} \\
\multicolumn{2}{c}{(a)} \\
\includegraphics[width=1.7in]{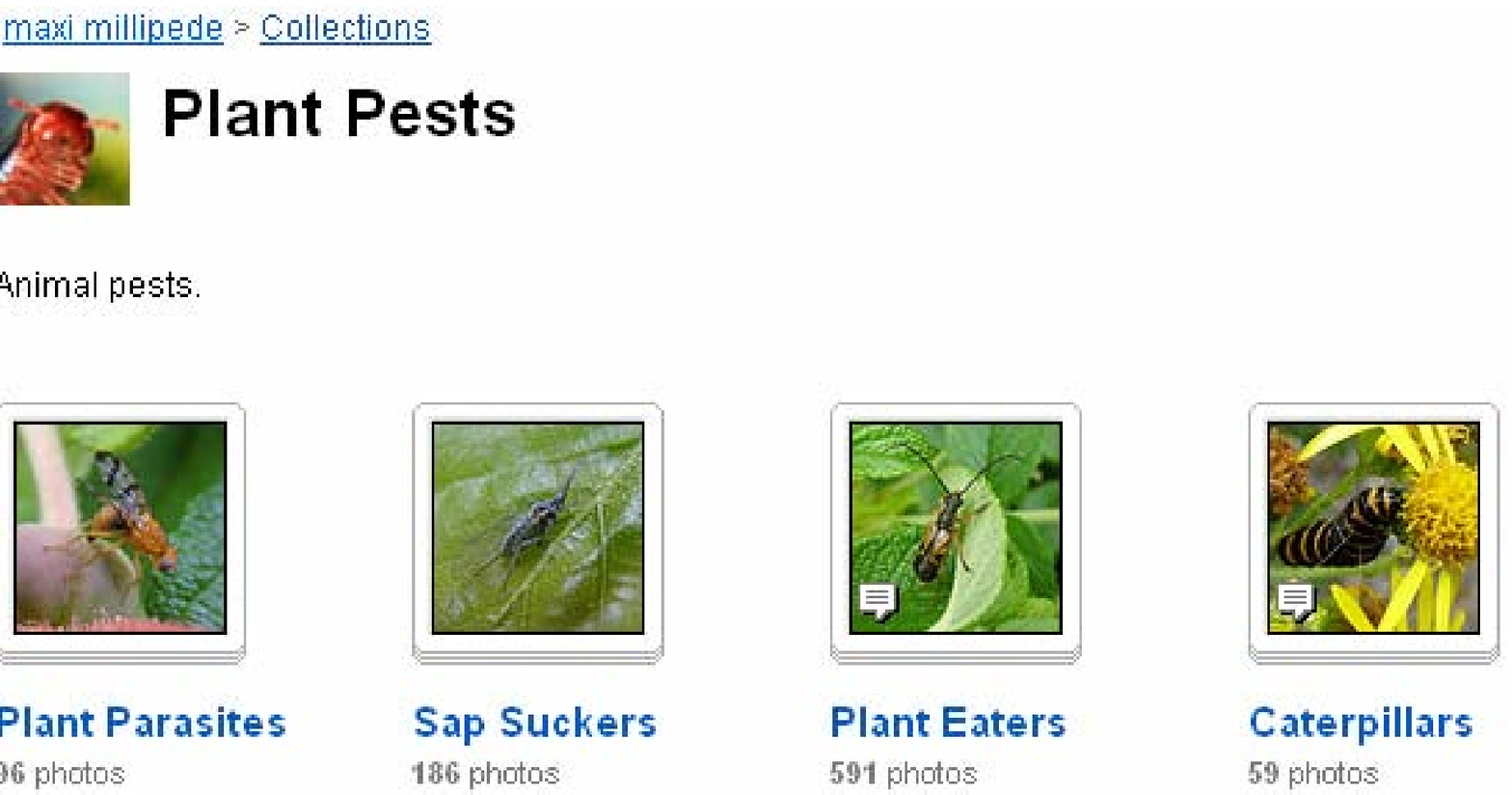} & \includegraphics[width=1.0in]{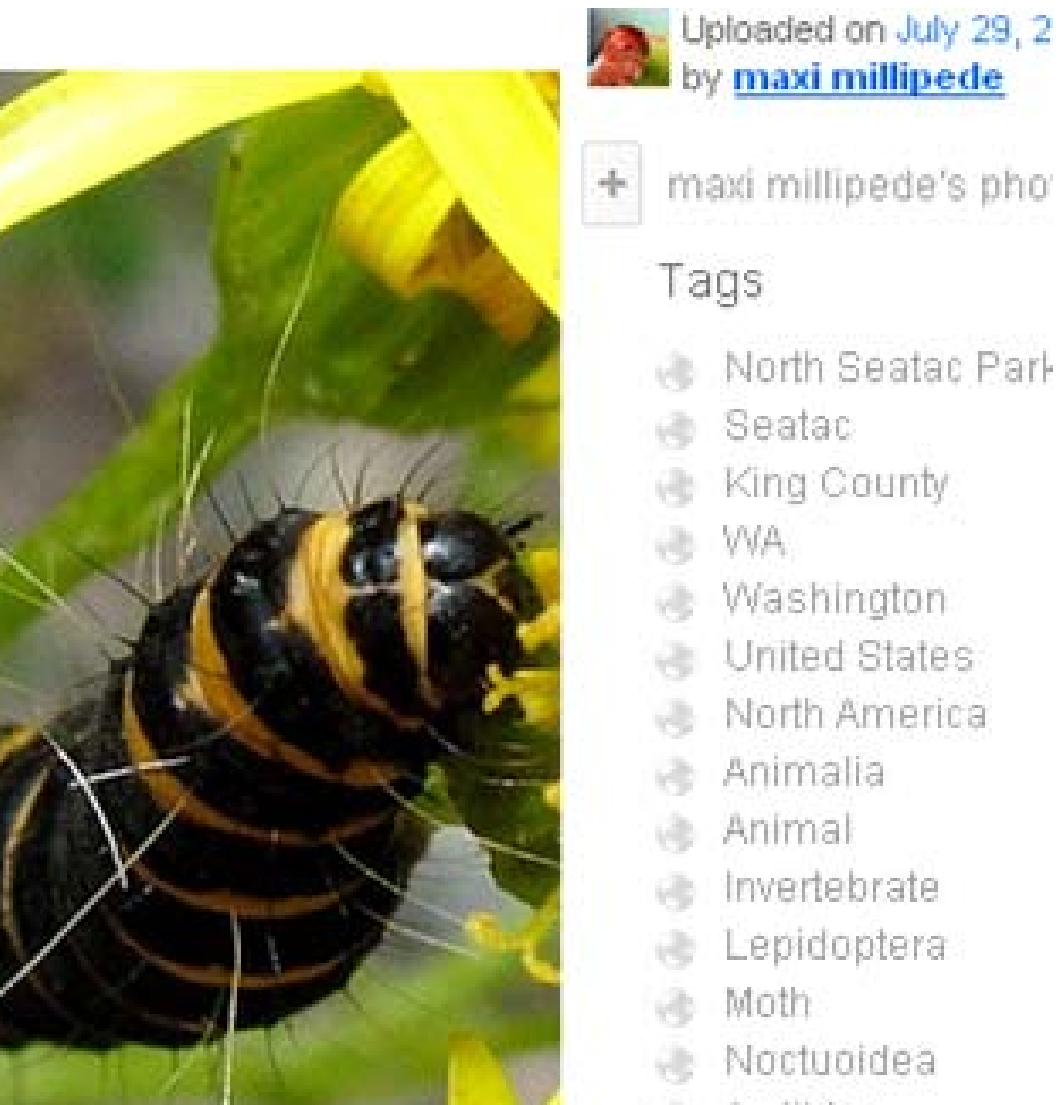} \\
(b) & (c) \\
\end{tabular}
  \caption{Personal hierarchies specified by a Flickr user. (a) Some of the collections created by the user and (b) sets associated with the \coll{Plant Pests} collection, and (c) tags associated with an image in the \set{Caterpillars} set.}\label{fig:collections}
\end{figure}

We briefly describe how this feature is implemented on the social photo-sharing site, \emph{Flickr} (\texttt{http://www.flickr.com}). Flickr  allows users to group their photos in album-like folders, called \emph{sets}. Users can also group sets into ``super'' albums, called \emph{collections}\remove{ (and related collections into new collections)}.\footnote{The collection feature is limited to paid ``pro'' users. Pro users can also create unlimited number of photo sets, while free membership limits a user to three sets.} Both sets and collections are named by the owner of the image. A photo can be part of multiple sets.

While Flickr does not enforce any specific rules about how to organize photos or how to name them, most users group ``similar'' or ``related'' photos into the same set and related sets into the same collection. Some users create multi-level hierarchies containing collections of collections, etc., but the vast majority of users create shallow hierarchies, consisting of collections and their constituent sets.
Figure~\ref{fig:collections}(a) shows some of the collections created by an avid naturalist on Flickr. These collections reflect the subjects she likes to photograph: \coll{Birds}, \coll{Mammals}, \coll{Plants}, \coll{Mushrooms \& Fungi}, \coll{Plant Pests}, \coll{Plant Diseases}, etc.  \figref{fig:collections}(b) shows sets of the \coll{Plant Pests} collection: \remove{``Mushrooms,'' ``Fungi, Puffballs, \& Shelf Fungi,'' and ``Molds and Rusts.''} \set{Plant Parasites}, \set{Sap Suckers}, \set{Plant Eaters}, and \set{Caterpillars}.
Each set contains one or more photos, which are tagged by the user.
For example, a photograph in the set \set{Caterpillars} (\figref{fig:collections}(c)), is annotated with multiple tags describing it:(\term{Animal}, \term{Lepidoptera}, \term{Moth}, \term{larva}, \term{Caterpillar}), its color (\term{Black and orange}), condition (\term{on Senecio}, \term{eating}), and location (\term{North Seatac Park}, \term{King County}, \term{WA}, \term{North America}).


\section{Challenges in Learning from \\Structured Metadata}
\label{sec:challenges}
Learning folksonomies from social metadata, specifically, from structured metadata, presents a number of challenges: \comment{We divide these challenges into categories.}


\begin{figure}[tb]
\begin{tabular}{c}
   \includegraphics[width=3.2in]{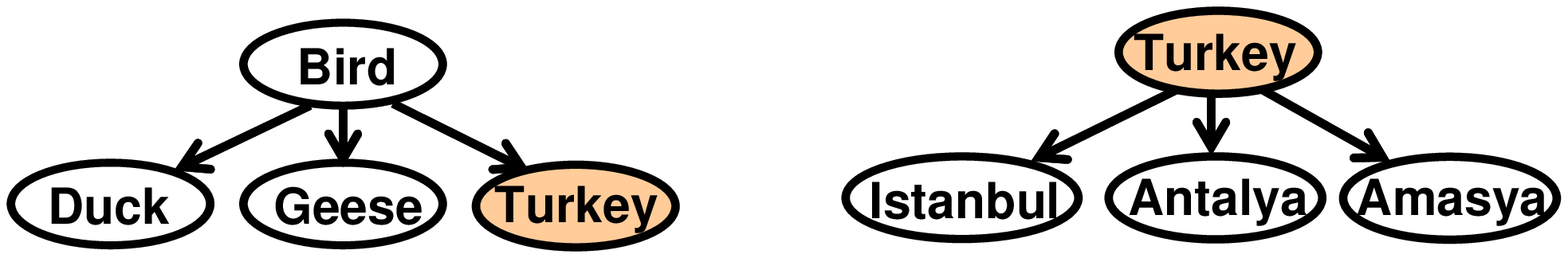} \\
  (a) Ambiguity \\
   \includegraphics[width=3.2in]{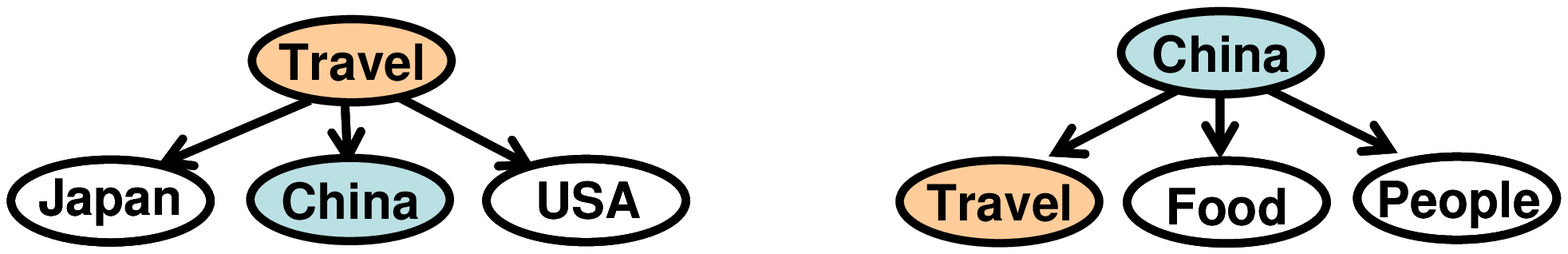} \\
	(b) Conflict\\
   \includegraphics[width=3.2in]{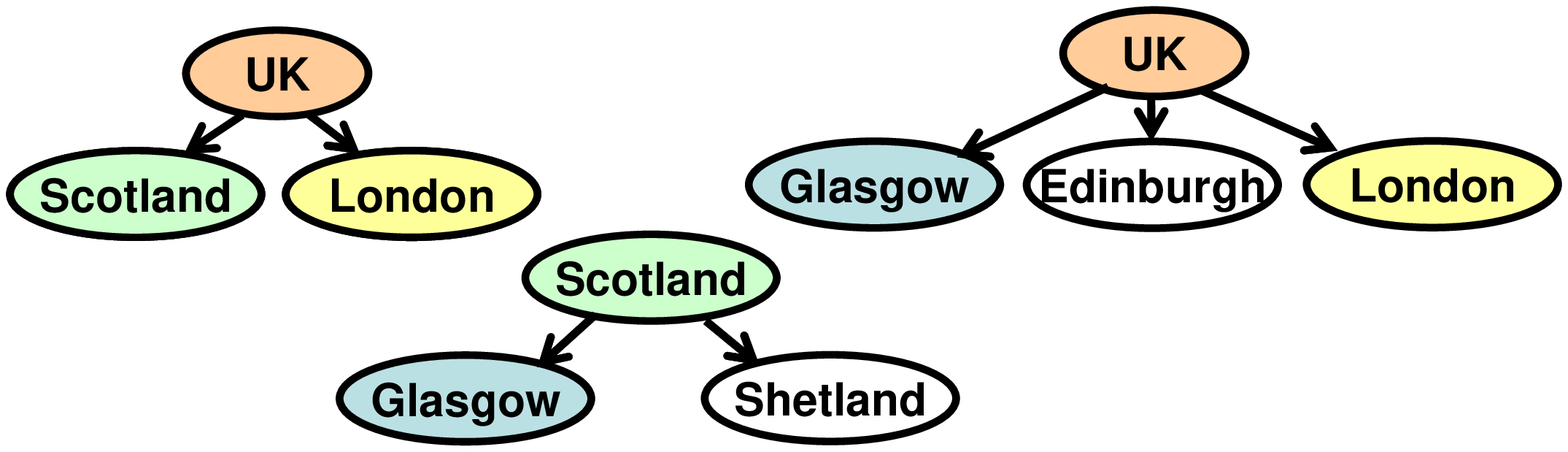} \\
	(c) Varying granularity
\end{tabular}
  \caption{Schematic diagrams of personal hierarchies created by Flickr users. (a) Ambiguity: the same term may have different meaning (``turkey'' can refer to a bird or a country). (b) Conflict: users' different organization schemes can be incompatible (\term{china} is a parent of \term{travel} in one hierarchy, but the other way around in another). (c) Granularity: users have different levels of expressiveness and specificity, and even mix different specificity levels within the same hierarchy (\term{Scotland} (country) and \term{London} (city) are both children of \term{UK}). Nodes are colored to aid visualization.}\label{fig:challenges}
\end{figure}

\subsection{Sparseness}
Social metadata is usually very sparse. Users provide $4$ --$7$ tags per bookmark on \emph{Delicious} in our data set and $3.74$ tags per photo on \emph{Flickr}~\cite{Rattenbury07}.
Sparseness is also manifested in the hierarchical organization created by an individual. In our Flickr data set, we found only $600$ out of $21,792$ users --- approximately $0.02$ percent --- who created multi-level (collections of collections) hierarchies. Most users define \emph{shallow} (single-level) hierarchies. Moreover, among these shallow hierarchies, few users organize content the same way.\remove{express ``similar'' organization.} For instance, of the $433$ users who created an \term{animal} collection, only a few created common child sets, such as \term{bird}, \term{cat}, \term{dog} or \term{insect}. In order to learn a rich and complete folksonomy, we have to aggregate social metadata from many different users.

\subsection{Noisy vocabulary}
\comment{
There are two main types of noise in social metadata: vocabulary and structure noise, which we explain later in this section.} Vocabulary noise has several sources. One common source is variations and errors in spelling. \remove{We can mitigate some of the noise by lowercasing and stemming terms in metadata.} Noise also arises from users' idiosyncratic naming conventions. While such names as \term{not sure}, \term{pleaseaddthistothethemecomppoll}, \term{mykid} may be meaningful to image owner and her narrow interest group, they are relatively meaningless to other users. 

\subsection{Ambiguity}
An individual tag is often ambiguous~\cite{Mathes04,Golder06}. For example, \term{jaguar} can be used to refer to a mammal or a luxury car. Similarly, terms that are used to name collections and sets can refer to different concepts. Consider the hierarchy in \figref{fig:challenges} (a), where \term{turkey} collection could be about a bird or a country. Similarly, \term{victoria} can either be a place in Canada or Australia. When combining metadata to learn common folksonomies, we need to be aware of its meaning. Structural and contextual information may help disambiguate metadata.

\subsection{Structural noise and conflicts} 
\remove{Conflicting metadata}
Like vocabulary noise, structural noise has a number of sources and can lead to inconsistent or conflicting structures.
Structural noise can arise as a result of variations in individuals' organization preferences. Suppose that, as shown in \figref{fig:challenges} (b), user $A$ organizes photos first by activity, creating a collection called \coll{travel}, and as part of this collection, a set called \set{china}, for photos of her travel in China. Meanwhile, user $B$ organizes photos by location first, creating a collection \coll{china}, with constituent sets \set{travel}, \set{people}, \set{food}, etc. In one hierarchy, therefore, \term{travel} is more general than \term{china}, and in the second hierarchy, it is the other way around. Sometimes conflicts are caused by vocabulary differences among individual users. For example, to some users \term{bug} is a ``{pest},'' a term broader than \term{insect}, while to others it is a subclass of \term{insect}. As a result, some users may express \term{bug} $\rightarrow$ \term{insect}, while the others express an inverse relation. \remove{Folksonomy integration should automatically resolve these types of conflicts.}
Another source of noise is variation in degree of expertise on a topic. Many users assemble images of spiders in a set called \set{spiders} and assign it to an \coll{insect} collection, while others correctly assign \set{spiders} to \coll{arachnid}.

\subsection{Varying granularity level} 
Differences in users' level of expertise and expressiveness \remove{granularity} may also lead to relatively imprecise metadata. Experts may use specific breed names to tag dog photos, while non-experts will simply use the tag \term{dog} to annotate them\cite{Golder06}. In addition, one user may organize photos first by country and then by city, while another organizes them by country, then subregion and then city, as shown in \figref{fig:challenges} (c). Combining data from these users potentially generates multiple paths from one concept to another.

\comment{
\begin{figure}
   \includegraphics[width=3in]{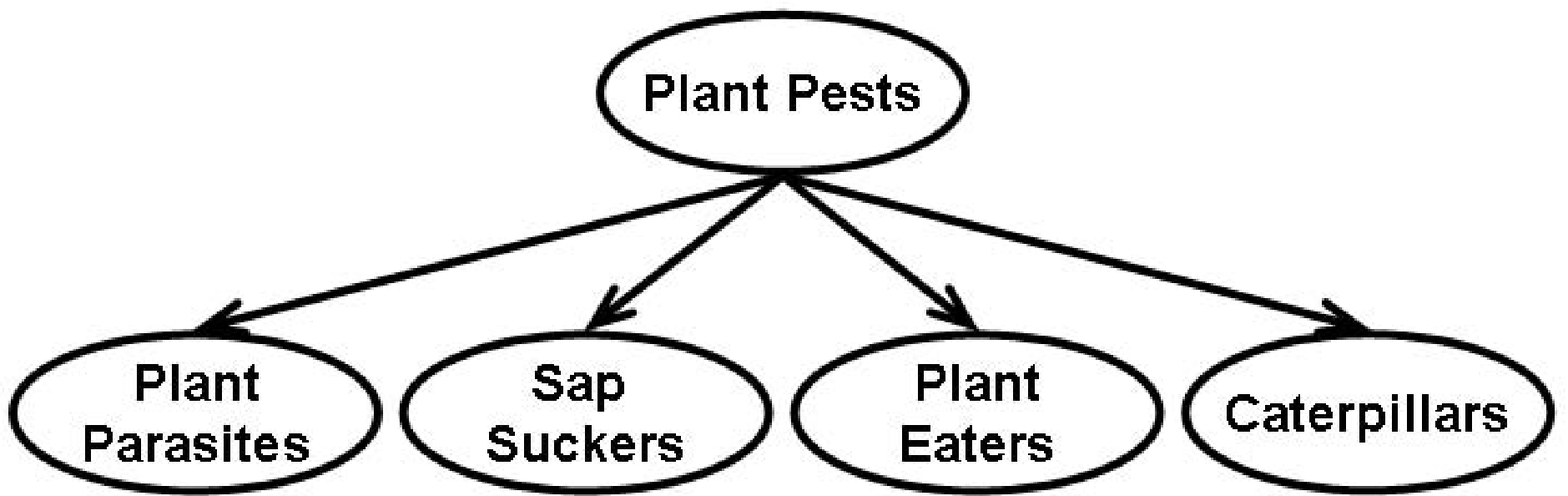}
  \caption{Shallow tree corresponding to user's top-level collection \coll{Plant Pests}}\label{fig:tree}
\end{figure}
}
\section{Learning Folksonomies from \\Structured Metadata}
\label{sec:approach}
We propose a simple, yet effective approach to combine many personal hierarchies into a global folksonomy that takes above challenges into account.
We define a personal hierarchy as a shallow {tree}, a \emph{sapling}, composed of a root node $r^{i}$ and its children, or leaf nodes
$\langle l^{i}_{1},..l^{i}_{j} \rangle$. The root node corresponds to a user's collection, and inherits its name, while the leaf nodes correspond to the collection's constituent sets and inherit their names.
Only a small number of users define multi-level hierarchies; for these, we decompose them and represent them as collections of saplings.  At the top level, we have a root node, which corresponds to the top-level collection, and its leaf nodes corresponding to the root's sets or collections. We then construct saplings that correspond to the leaf nodes, which are collections, and so on. We assume that hierarchical relations between a root and its children, $r^{i} \rightarrow l^{i}_{j}$, specify broader-narrower relations.  \comment{Figure~\ref{fig:tree} shows the sapling constructed from the collection \coll{Plant Pests} created by the user in \figref{fig:collections}. In this case, \term{Plant Pests} is a broader concept than \term{Caterpillars} or \term{Sap Suckers}.} Hence, the sapling in \figref{fig:collections} (b) is  \term{Plant Pests} $\rightarrow$ \{\term{Plant Parasites}, \term{Sap Suckers}, \term{Plant Eaters}, \term{Caterpillars} \}.

%

In addition to hierarchical structure, each sapling carries information derived from tags.
On Flickr, users attach tags only to photos; therefore, the tag statistics of a sapling's leaf (set) are aggregated from that set's constituent photos. Tag statistics are then propagated from the leaves to the parent node. In
our example, \set{Plant Parasites} aggregates tag statistics from all photos in this set, and its parent \coll{Plant Pests} contains tag statistics accumulated from all photos in \set{Plant Parasites} and its siblings.
We define a tag statistic of node $x$ as $\tau_{x} := \{(t_{1},f_{t_{1}}), (t_{2},f_{t_{2}}), \cdots (t_{k},f_{t_{k}})\} $, where $t_{k}$ and $f_{t_{k}}$ are \emph{tag} and its frequency respectively. Hence, $\tau_{r^{i}}$ is aggregated from all $\tau_{l^{i}_{j}}$s.


\comment{
We assume that each node in a sapling refers to a concept about the world. We expect, therefore, that the node \term{Plant Parasites}
refers to images of plant parasites. Given a collection of saplings, specified by many different users, our goal is to aggregate them into a common, denser and deeper tree. Before describing the approach to integrate saplings, we first briefly describe data preprocessing step, in order to address the sparseness and noise challenges listed above.}
%
Given a collection of saplings, specified by many different users, our goal is to aggregate them into a common, denser and deeper tree. Before describing our approach, we first briefly describe data preprocessing steps that address the sparseness and noise challenges listed above.

\subsection{Data Preprocessing}
\label{sec:preproc}
We extract terms representing
concepts
from collection and set names. We found that users often combine two or more concepts within a single name, \remove{ by using words and special characters to join different concepts,} e.g., ``Dragonflies/Dam\-sel\-flies'', ``Mushrooms \& Fungi'', ``Moth at Night.''
Terms can be joined by bridging words that include prepositions ``at'', ``of'', ``in,'' and conjunctions ``and'' and ``or,''  or special characters, such as `\&', `$<$', `$>$', `:', `/'. We start by tokenizing collection and set names on these words and characters. We do not tokenize on white spaces to avoid breaking up terms like ``South Africa.'' We remove terms composed only of non-alphanumeric characters and frequently-used uninformative words, \emph{e.g.}, ``me'' and ``myself.'' We then normalize all terms by lowercasing them.

After tokenization, a set or collection name may be split into multiple terms, which we expand into leaves. Suppose a user created a collection
\coll{animal} containing a set \set{cats and dogs}. \remove{Since ``cats and dogs'' is determined not to be a composite name by the likelihood test, the sapling we get after} After tokenization we get the sapling \term{animal} $\rightarrow$ \{\term{cats}, \term{dogs}\}.
However, if the root node is determined to have a composite name,
we ignore the entire sapling because we do not know which parent concepts correspond to which child concepts.

\subsection{Relational Clustering of Structured Metadata}
In order to learn a folksonomy, we need to aggregate saplings \emph{both} horizontally and vertically. By horizontal aggregation, we mean merging saplings with similar roots, which expands the breadth of the learned tree by adding leaves to the root. By vertical aggregation, we mean merging one sapling's leaf to the root of another, extending the depth of the learned tree. The approach we use exploits contextual information from neighbors in addition to local features to determine which saplings to merge. The approach is similar to relational clustering\cite{EntityResoluteGetoor07} and its basic element is the similarity measure between a pair of nodes.

We define a similarity measure \remove{between nodes in different saplings,} which combines heterogeneous evidence available in the structured social metadata, and is a combination of \emph{local similarity} and \emph{structural similarity}. The local similarity between nodes $a$ and $b$, $localSim(a,b)$, is based on the
intrinsic features of $a$ and $b$, such as their names and tag distributions. The structural similarity, $structSim(a,b)$ is based on features of neighboring nodes. If $a$ is a root of a sapling, its neighboring nodes are all of its children. If $a$ is a leaf node, the neighboring nodes are its parent and siblings.
The similarity between nodes $a$ and $b$ is:
\begin{eqnarray}
\label{eq:grandfunction}
nodesim(a,b) &=& (1-\alpha) \times localSim(a,b) \\ \nonumber
						 &+& \alpha \times structSim(a,b),
\end{eqnarray}
\noindent where $0 \le \alpha \le 1$ is a weight for adjusting contributions from $localSim(,)$ and $structSim(,)$. We judge whether two nodes are similar if the similarity is greater than the threshold, $\tau$.




\comment{
Computing the similarity measure for all pairs of sapling nodes in the corpus is impractical, even considering local or structural similarity only. Therefore, we simplify the problem in two ways.
First, we retain only the top
$K$ most-frequent tags for each node. This will reduce the time complexity when comparing tag statistics for each node pair. Second, we only compare sapling nodes if they share (stemmed) name.
This reduces the total number of pairs which need to be compared.
}

\subsubsection{Local Similarity}

The local similarity of nodes $a$ and $b$ is composed of (1) name similarity and (2) tag distribution similarity.
\comment{
Since, as
explained above, we only consider merging nodes with the same name, name similarity is already taken into account. Consequently, the
remaining decision is based on similarity of tag distribution, $tagSim(a,b)$.
}
Name similarity can be any string similarity metric, which returns a value ranging from $0$ to $1$.
Tag similarity, $tagSim(,)$, can be any function for measuring the similarity of distributions. Because of the sparseness of the data, and to make the computation fast, we use a simple function which counts the number of common tags, $n$, in the top $\texttt{K}$ tags of $a$ and $b$; it returns $1$ if this number is equal or greater than $J$, else it returns $\frac{n}{\texttt{J}}$. 
\remove{
Tag similarity can be a simple function that, e.g., checks the number of common tags, say $n$, in the top $\texttt{K}$ tags of $a$ and $b$, and returns $1$ if this number is equal or greater than $\texttt{J}$; otherwise, it returns $\frac{n}{\texttt{J}}$. \comment{We use a logistic function to smooth the step function.}} Local similary is a weighted combination of name and tag similarities:
\begin{eqnarray}
\label{eq:pracLocal}
localSim(a,b) &=&  \beta \times nameSim(a,b) \\ \nonumber &+& (1-\beta) \times tagSim(a,b)).
\end{eqnarray}

Tag similarity helps address the \emph{ambiguity} challenge described in \secref{sec:challenges}.
For example, the top tags of the node \term{turkey} that refers to a bird include ``bird'', ``beak'', ``feed'', while the top tags of \term{turkey} that refers to the country include different terms about places \comment{and attractions} within the country.

\subsubsection{Structural Similarity}
Structural similarity between two nodes depends on position of nodes within their saplings. We define two versions: $structSimRR(,)$ which computes structural similarity between two root nodes (root-to-root similarity), and $structSimLR(,)$ which evaluates structural similarity between a root of one sapling and the leaf of another (leaf-to-root similarity).
\paragraph*{Root-to-Root similarity}
Two saplings $A$ and $B$ are likely to describe the same concept if their root nodes $r^{A}$ and $r^{B}$ 
\remove{share the same (stemmed)}
have a similar name and some of their leaf nodes also have similar \remove{the same} names.
In this case, there is no need to compute $tagSim(,)$ of these leaf nodes. We define the normalized common leaves factor, $\mathds{CL}$, as $\frac{1}{Z}\sum_{i,j} \delta(name(l^{A}_{i}),name(l^{B}_{j}))$, where $\delta(.,.)$ returns $1$ if the both arguments are exactly the same; otherwise, it returns $0$; $name(l^{A}_{i})$ is a function that returns the name of a leaf node $l^{A}_{i}$ of sapling $A$. $Z$ is a normalizing constant, which is described in greater detail later. Structural similarity between two root nodes is then defined as follows:
\begin{eqnarray}
\label{eq:pracStructRR}
structSimRR(r^{A},r^{B})
= \mathds{CL}&+&(1-\mathds{CL}) \\ \nonumber
      &\times& tagSim(\mathds{\grave{L}}^{A}_{tag},\mathds{\grave{L}}^{B}_{tag}),
\comment{
max \left\{
\begin{array}{l l}
    \frac{1}{Z}\sum_{i,j} \delta(name(l^{A}_{i}),name(l^{B}_{j})) \\
    tagsim(\mathds{\grave{L}}^{A}_{tag},\mathds{\grave{L}}^{B}_{tag})\\
\end{array} \right.
}
\end{eqnarray}
\noindent
\comment{
where $\delta(.,.)$ returns $1$ if the both arguments are exactly
the same; otherwise, it returns $0$.
$name(l^{A}_{i})$ is a function
that returns the name of a leaf node $l^{A}_{i}$ of sapling $A$, and
}
\noindent where $\mathds{\grave{L}}^{A}_{tag}$ is an aggregation of tag distributions
of all $l^{A}_{i}$, at which $name(l^{A}_{i}) \neq name(l^{B}_{j})$
for any leaf node $l^{B}_{j}$ of the sapling $B$. 
\comment{$\mathds{\grave{L}}^{B}_{tag}$ is defined in the similar way.} From \eqrefx{eq:pracStructRR},
we compute similarity based on: (1) how many of their
children have common name (they match); (2) the tag
distribution similarity of those that do not have the same name. The
second term is an optimistic estimate that child nodes of the two
saplings refer to the same concept while having different names. \comment{We
simply use $max(,)$ function to choose the higher score.}

The normalization coefficient $Z=min(|l^X|,|l^Y|)$, where
$|l^X|$ is a number of child nodes of $X$. We use $min(,)$ instead of
union. The reason is that saplings aggregated from many
small saplings will contain a large number of child nodes. When merging with
a relatively small sapling, the fraction of common nodes may be very
low compared to total number of child nodes. Hence, the
normalization coefficient with the union ($Z = union(l^X,l^Y)$), as
defined in Jaccard similarity, results in overly penalizing small saplings.  $min(,)$, on the other hand, seems to correctly consider
the proportion of children of the smaller sapling that overlap with
the larger sapling.

\comment{
Evidence from leaf nodes can also help disambiguate saplings. For example, saplings about \term{victoria}, which refer to the place in Canada, would have different children than saplings about \term{victoria} in Australia, and would not be merged together.
}


When we decide that roots
$r^{A}$ and $r^{B}$ are similar, we merge saplings $A$ and $B$ with the
$merge\-By\-Root(A,B)$ operation. This operation creates a new sapling, $M$, which combines structures and tag statistics of $A$ and $B$. In particular, the tag statistics of the root of $M$ is a combination of those from $r^{A}$ and $r^{B}$. The leaves of $M$, $l^{M}$, are a union of $l^{A}$ and $l^{B}$. If there are leaves from $A$ and $B$ that share a name, their tag statistics will be combined and attached to the corresponding leaf in $M$.   

\comment{
\begin{itemize}
\item Let $r^{M}$ be the root of the newly merged sapling $M$, having $stem(r^{M}) = stem(r^{A})$
\item $\tau_{r^{M}} \leftarrow \tau_{r^{A}} \oplus \tau_{r^{B}}$,
\item For each $n$ in $\{stem(l^{A}) \cup stem(l^{B})\}$,
\begin{itemize}
\item		Let $l^{M}_{k}$ be a leaf of $M$, having $stem(l^{M}_{k}) = n$,				
\item   if $stem(l^{A}_{i}) =  stem(l^{B}_{j})$ and $stem(l^{A}_{i}) = n$, \\ $\tau_{l^{M}_{k}} \leftarrow \tau_{l^{A}_{i}} \oplus \tau_{l^{B}_{j}}$;
%

\item		otherwise, $\tau_{l^{M}_{k}} \leftarrow \tau_{l^{A}_{i}}$ or $\tau_{l^{B}_{j}}$

\end{itemize}
\end{itemize}

\noindent We define $\oplus$ as a function which combines tag frequencies  ($\tau$s) from two sets. In words, the merged sapling $M$ combines all leaf nodes and tag statistics from $A$ and $B$.} 

The width of the newly merged sapling will increase as more saplings are merged. Also, since we simply merge leaf nodes with similar names, and their roots also have similar names, leaf-to-leaf structural similarity $structSimLL(,)$ is not required. This operation addresses the \emph{sparseness} challenge mentioned in \secref{sec:challenges}.
\paragraph*{Root-to-Leaf similarity}

Merging the root node of one sampling with the leaf node of another sapling extends the depth of the learned
folksonomy. Since we consider a pair of nodes with different roles,
their neighboring nodes also have different roles. This would appear
to make them structurally incompatible.  However, in many cases,
some overlap between siblings of one sapling and children of
another sapling exists. Formally, suppose that we are
considering similarity between leaf $l^{A}_{i}$ of sapling $A$ and
root $r^{B}$ of sapling $B$. There might be some $l^{A}_{k \neq i}$ of
$A$ similar to $l^{B}_{j}$ of $B$.  Consider \figref{fig:challenges}
(c). Suppose that we have already merged \term{uk} saplings. Now, there
are two saplings \term{uk} $\rightarrow$ \{\term{scotland},
\term{glasgow}, \term{edinburgh}, \term{london}\} and \term{scotland}
$\rightarrow$ \{\term{glasgow}, \term{shetland}\}, and we would like
to merge the two \term{scotland}s. Since both \term{uk} and
\term{scotland} saplings have \term{glasgow} in common, and the user
placed \term{glasgow} under \term{uk} instead of \term{scotland}, this
\emph{shortcut} contributes to the similarity between \term{scotland}
nodes. The structural similarity between leaf and root nodes that
takes this type of shortcut into consideraion is:
\begin{equation}
\label{eq:pracStructLR}
structSimLR(l^{A}_{i},r^{B}) = structSimRR(r^{A},r^{B}).
\comment{
structSimLR(l^{A}_{i},r^{B}) = \left\{
\begin{array}{l l}
	structSimRR(r^{A},r^{B}) & \mbox{if \emph{C1}} \\
  0.5 & \mbox{otherwise}

\end{array} \right.
}
\end{equation}

\comment{The first condition, \emph{C1}, is the case when a shortcut is created by merging, or formally, $\{stem(l^{A})\} \cap \{stem(l^{B})\} \neq \O$.}
Specifically, this is simply the root-to-root structural similarity of $r^{A}$ and $r^{B}$, which measures overlap between siblings of $l^{A}_{i}$ and children of $r^{B}$. For the case when there is no shortcut, the similarity from this part will be dropped out; hence, the \eqrefx{eq:grandfunction} will only be based on the local similarity.


\subsection{SAP: Growing a Tree by Merging Saplings}

We describe \algname{sap} algorithm, which uses operations defined above to incrementally grow a deeper,
bushier tree by merging saplings created by different users. In order
to learn a folksonomy corresponding to some concept, we start by
providing a seed term, the name of that concept. The seed term will be
the root of the learned tree.
We cluster individual saplings whose
roots have the same name as the seed by using the similarity measures
\eqrefx{eq:grandfunction}, \eqrefx{eq:pracLocal} and \eqrefx{eq:pracStructRR} to identify
similar saplings. Saplings within the same cluster are merged into a
bigger sapling using the $mergeByRoot(,)$ operation. Each merged
sapling corresponds to a different sense of the seed term.

Next, we select one of the merged saplings as the starting point for
growing the folksonomy for that concept. For each leaf of the initial
sapling, we use the leaf name to retrieve all other saplings
whose roots are similar to the name. We then merge saplings
corresponding to different senses of this term as described above. The
merged sapling whose root is most similar to the leaf (using
similarity measures \eqrefx{eq:grandfunction}, \eqrefx{eq:pracLocal} and
\eqrefx{eq:pracStructLR}), is then linked to the leaf. In the case that
several saplings match the leaf, we merge all of them together
before linking.
Clustering saplings into different senses, and then merging relevant
saplings to the leaves of the tree proceeds incrementally until some
threshold is reached.

Suppose we start with saplings shown in
\figref{fig:challenges}(c), and the seed term is \term{uk}. The process
will first cluster \term{uk} saplings. Suppose, for illustrative
purposes, that there is only one sense of \term{uk}, resulting in a
single sapling with root \term{uk}. Next, the procedure selects one of
the unlinked leaves, say \term{glasgow}, to work on. All saplings with
root \term{glasgow} will be clustered, and the merged \term{glasgow}
sapling that is sufficiently similar to the \term{glasgow} leaf of the
\term{uk} sapling will then be linked to it at the leaf, and so on.


\paragraph*{Handling Shortcuts}

Attaching a sapling $A$ to the learned tree $F$ can result in
structural inconsistencies in $F$. One type of
inconsistency is a \emph{shortcut}, which arises when a
leaf of $A$ is similar to a leaf of $F$.
%
%
In the illustration above, attaching the \term{scotland} sapling to the
\term{uk} tree will generate a shortcut, or two possible paths from
\term{uk} to \term{glasgow} ($r^{uk} \rightarrow l^{uk}_{glassgow}$
and $r^{uk} \rightarrow l^{uk}_{scotland} \rightarrow
l^{scotland}_{glasgow}$). Ideally, we would drop the shorter path and
keep the longer one which captures more specific knowledge.

There are cases where the decision to drop the shorter path cannot be
made immediately. Suppose we have \term{uk} $\rightarrow$
\{\term{london}, \term{england}, \term{scotland}\} as the current
learned tree, and are about to attach \term{london} $\rightarrow$
\{\term{british museum}, \term{dockland}, \term{england}\} to
it. Unfortunately, some users placed \term{england} under
\term{london}, and attaching this sapling will create a shortcut to
\term{england}. The decision to eliminate the shorter path to
\term{england} cannot be made at this point, since we have no
information about
whether attaching the
\term{england} sapling will also create a shortcut to \term{london}
from the root (\term{uk}). We have to postpone this decision until we
retrieve all relevant saplings that can be attached to the present
leaf ($l^{uk}_{london}$) and its siblings ($l^{uk}_{england}$ and
$l^{uk}_{scotland}$).

\begin{figure}
   \includegraphics[width=3.3in]{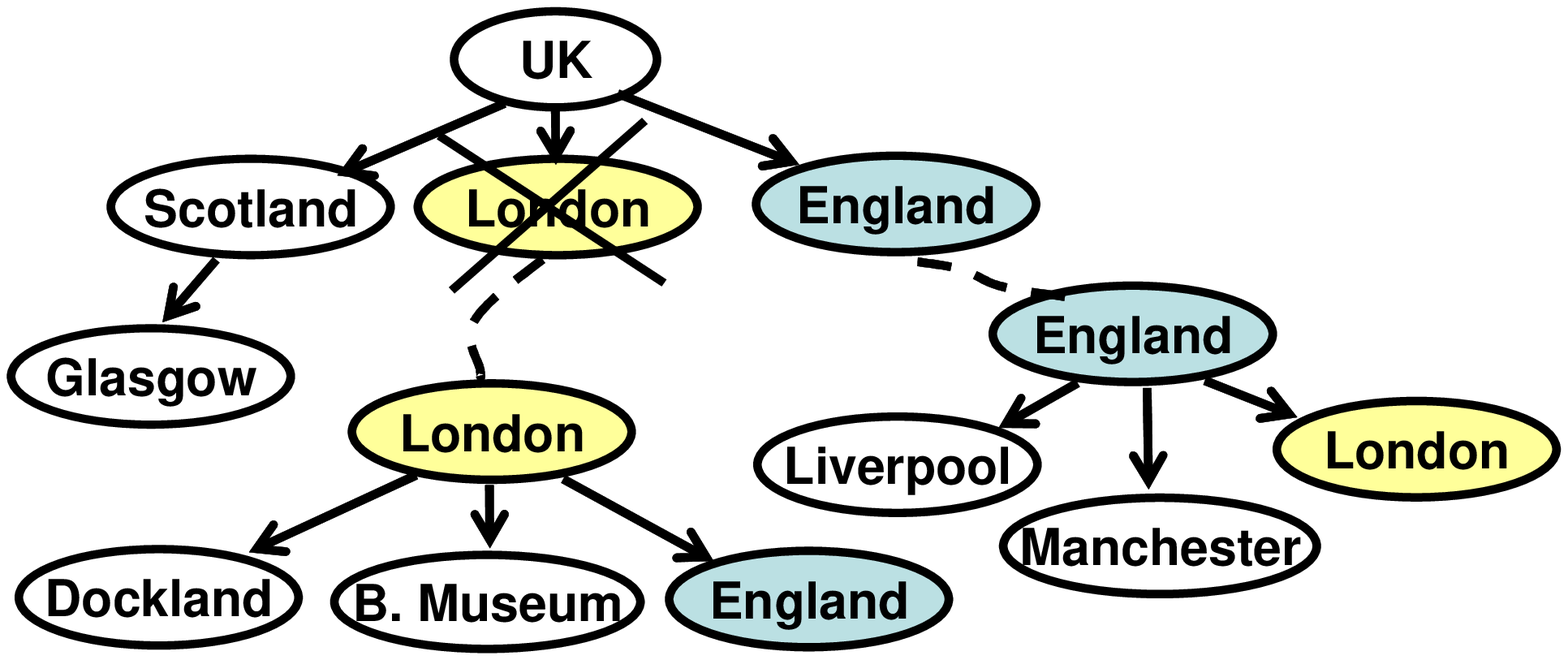}
  \caption{Appearance of mutual shortcuts between London and England when merging London and England saplings. To resolve them, we compare the similarity between UK-London and UK-England sapling pairs. Since England sapling is closer to UK than London sapling, we simply attach England sapling to the tree; while ignoring London leaf under UK.}\label{fig:mutualshortcut}
\end{figure}

Suppose that $l^{uk}_{england}$ does match the root of sapling
\term{eng\-land} $\rightarrow$ \{\term{london}, \term{manchester},
\term{liverpool}\}. Mutual shortcuts to \term{england} and
\term{london} would undesirably appear once all the saplings are
attached to the tree. Hence, the decision to drop $l^{uk}_{england}$
or $l^{uk}_{london}$ must be made. We base the decision on similarity.
Intuitively, a sapling that is more similar, or ``closer,'' to $r^{uk}$ should be linked to the tree.
Formally, the node to be kept is $l^{uk}_{\hat{x}}$, where $\hat{x} = argmax_{x}\{nodesim(r^{uk},r^{x})\}$ and $x=\{england, london\}$, while the other will be dropped. This is illustrated in \figref{fig:mutualshortcut}.

\paragraph*{Handling Loops}
Attaching a sapling to a leaf of the learned tree may result in
another undesirable structure, a \emph{loop}. Suppose
that we are about to attach a sapling $A$ to the leaf $l^{F}_{i}$ of $F$.
A loop will appear if there exists a leaf $l^{A}_{j}$ of $A$ with the
same name as some node in the path from root to
$l^{F}_{i}$  in $F$.
In order to make the learned tree
consistent, we must remove $l^{A}_{j}$
before attaching the sapling. For instance, suppose we decide to attach \term{london} sapling to the \term{england} sapling in \figref{fig:mutualshortcut} at its \term{london} node, we have to remove \term{england} node of \term{london} sapling first. 

In some cases, loops indicate synonymous concepts.  In our data set,
we found that there are users who specify the relation \term{animal}
$\rightarrow$ \term{fauna}, and those who specify the inverse
\term{fauna} $\rightarrow$ \term{animal}. Since \term{animal} and
\term{fauna} have similar meaning, we hypothesize that this conflict appears because of variations in users' expertise and
categorization preferences.

To determine whether a loop is caused by a synonym, we check the
similarity between $r^{A}$ and $r^{F}$. If it is high enough, we
simply remove $l^{F}_{j}$ from $F$, for which $name(l^{F}_{j}) =
name(l^{A}_{j})$; then, merge $r^{A}$ and $r^{F}$. The similarity
measure is based on \eqrefx{eq:grandfunction}. More
stringent criteria are required since $r^{A}$ and $r^{F}$ have
different names. Specifically, we modify $tagSim(X,Y)$ to
$tagSim^{syn}(X,Y)$, which instead evaluates $\frac{|\tau_{X} \cap
\tau_{Y}|}{min(|\tau_{X}|,|\tau_{Y}|)}$, and modify $structSim(X,Y)$
to $struct\-Sim^{syn}(X,Y)$, which only evaluates $\frac{1}{Z}\sum_{i,j}
\delta(name(l^{X}_{i})$,\\ $name(l^{Y}_{j}))$.

\paragraph*{Mitigating Noisy Vocabularies} As mentioned in \secref{sec:challenges}, noisy nodes appear from idiosyncratic vocabularies, used by a small number of users. For a certain merged sapling, we can identify these nodes by the number of users who specified them. Specifically, we use $1\%$ of the number of all users who ``contribute'' to this merged sapling as the threshold. We then remove leaves of the sapling, that are specified by fewer number of users than the threshold.

\paragraph*{Managing Complexity} 

Computing the similarity measure for all pairs of saplings in the corpus is impractical, even considering local or structural similarity only. We address this scalability issue in two ways.
\comment{
First, we retain only the top
$K$ most-frequent tags for each node. This will reduce the time complexity when comparing tag statistics for each node pair.}  
First, we only compare sapling nodes if they share the same (stemmed) name.
This reduces the total number of pairs which need to be compared and eliminates the need to compute $nameSim(,)$ in \eqrefx{eq:pracLocal}. Second, we apply the blocking approach \cite{MongeElkan97} for efficiently computing similarity and merging sapling roots. The basic idea behind this approach is to first use a cheap similarity measure to
``roughly'' group similar items. We can then thoroughly compute item similarities and merge them within each ``roughly similar'' group by using the more computationally expensive similarity measure. We assume that items judged to be dissimilar by the cheap measure will also be dissimilar when evaluated by the more expensive measure. Since the approach applies the expensive measure to a much smaller set of items, it reduces the time complexity of the clustering method.

In our case, we compute an inexpensive similarity measure based on the most frequent tags. Specifically, we map the top tags to some integer code, which can be cheaply sorted by any database. Subsequently, we use the database to sort saplings by their codes, moving roughly similar saplings to neighboring rows. The process begins by scanning sorted saplings in the database table on a sapling by sapling basis. If the presently scanned sapling has not been merged with some other sapling, we add this sapling to the top of the queue. If the present sapling does belong to some merged sapling, we check if this sapling is also similar to some other merged saplings in the queue. We use \eqrefx{eq:grandfunction}, \eqrefx{eq:pracLocal} and \eqrefx{eq:pracStructRR} to evaluate their similarity. If they are similar enough, we will merge them together into a new merged sapling; then add it to the top of the queue. The scanning is performed repeatedly until the number of merged saplings no longer changes.
\subsection{Complexity Analysis}
Here we sketch the computational complexity of \algname{sap}. Basically, \algname{sap} can be decomposed into 2 different parts: (1) root-to-root merging, which expands folksonomies' width; (2) leaf-to-root merging, which extends folksonomies' depth. These two parts are loosely dependent, i.e., one can cluster all saplings into different senses; then ``vertically'' merge the root of one sapling sense with a leaf of the other. Since we use blocking  and only cluster saplings with the same stemmed names, the computational complexity depends on (1) the number of unique stemmed names in the data set; (2) the average number of saplings that share a name. Let $N$ and $M$ be the number of nodes and the number of unique stemmed names in the data set respectively. Hence, for each stem, there are $\frac{N}{M}$ nodes to be compared on average. We use database to first roughly sort saplings, which generally requires $O(\frac{N}{M}log(\frac{N}{M}))$. After saplings are sorted, they are scanned and merged. This is repeatedly, say in $i$ iterations, until the number of clusters no longer changes, which requires $O(i\times\frac{N}{M})$. In all, the complexity of the first part is $O(Nlog(\frac{N}{M}) + iN)$. Empirically, the number of clusters converges in 2-3 iterations on average.

Let $b$ and $d$ be the branching factor and the depth of the tree we want to produce. In addition, suppose that there are $s$ sapling senses for each stemmed name on average. Since we have to traverse each inner node of the tree to attach relevant sapling senses, and for each of these nodes we need to compare the similarity to all sapling senses with similar root names, this requires $O(s\times b^{d})$.

Our earlier work, \algname{sig}~\cite{www09folksonomies}, which is described in more detail in \secref{sec:results}, only considered the best path from a root to a given leaf of the tree, and required enumerating all possible paths between them. In the best case, when there are no shortcuts or loops in the data set, the number of paths from the root to all leaves of a given tree is equal to the number of the leaves, and that only requires $O(b^{d}+b^{d-1})$ to check whether each edge should be included. In the worst case, when shortcuts appear to all node pairs, we would need $O(\binom{d+1}{2}\times b^{d})$ to check all possible edges. Moreover, we also need to enumerate all possible paths for the root to all leaves of the tree, which requires $O(1+\sum_{e=1:d-1}\binom{d-1}{e})$ per root-to-leaf pair. Hence, we expect our approach to scale better than the previous one as the depth of the output tree increases and when there are many shortcuts.   




\section{Empirical Validation}
\label{sec:results}

\begin{figure}[tbh]
\begin{tabular}{c}
   \includegraphics[width=3.3in]{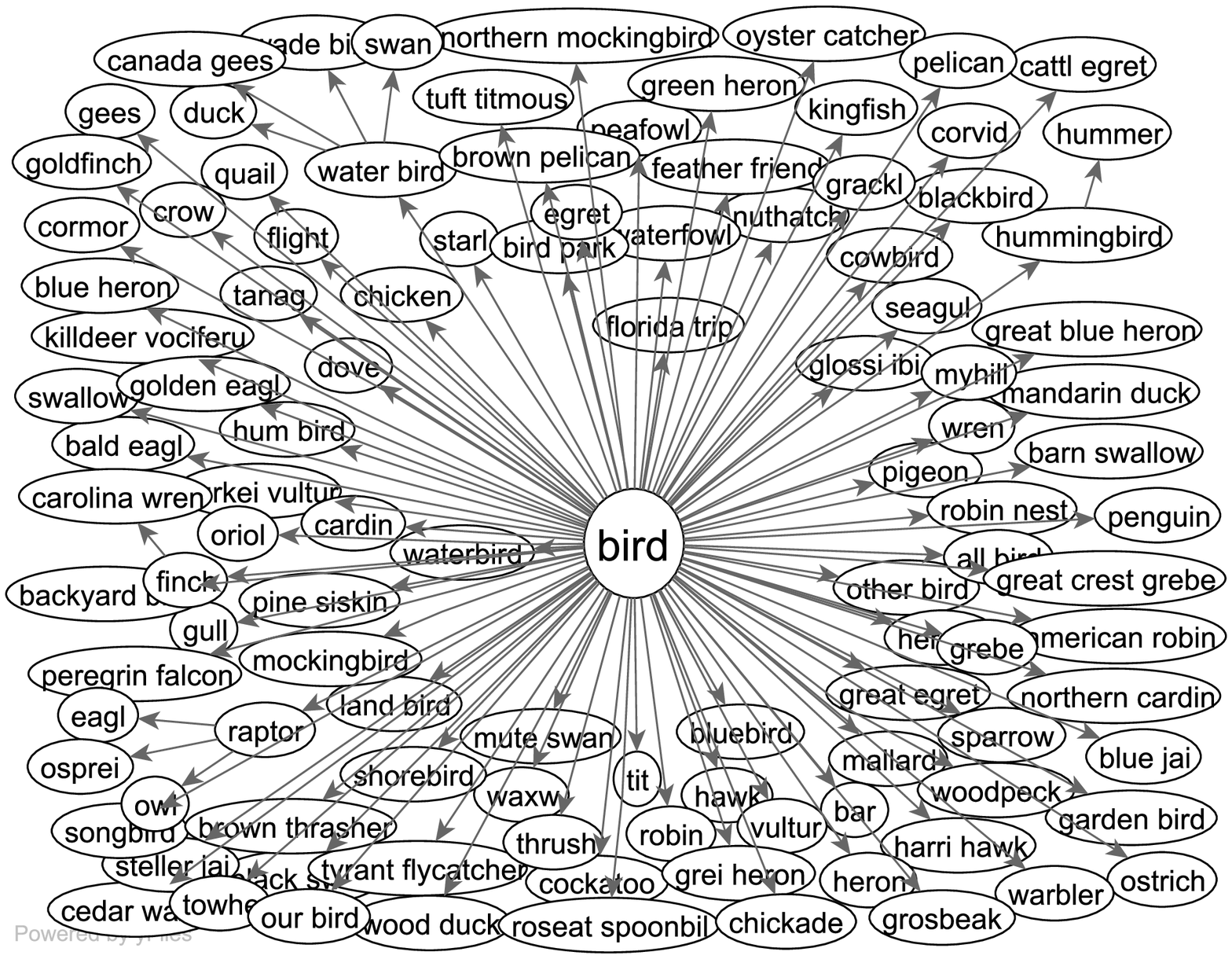} \\ 
   \includegraphics[width=2.8in]{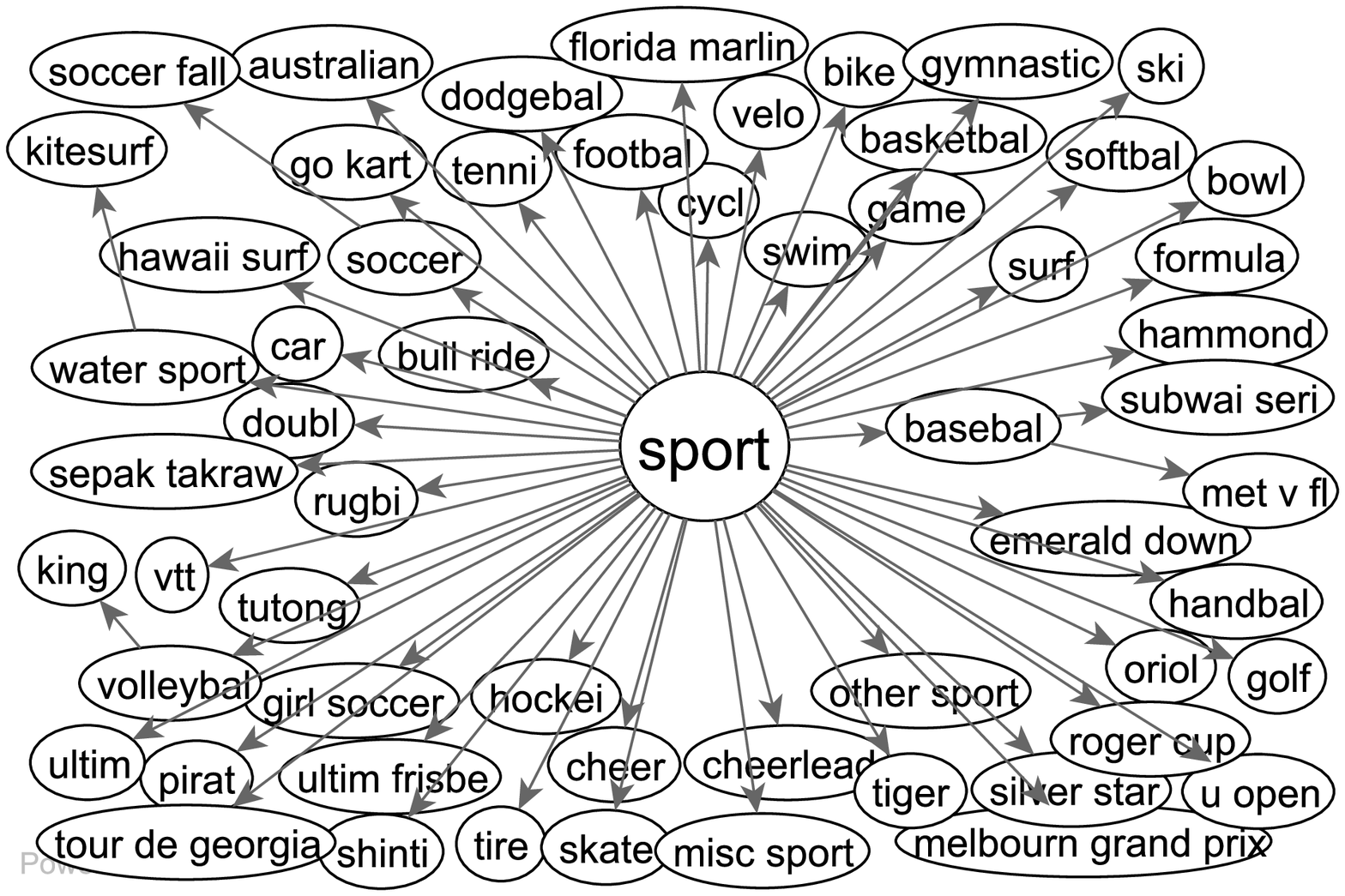} \\
\end{tabular}
  \caption{Folksonomies learned for \term{bird} and \term{sport}}\label{fig:trees}
\end{figure}

We constructed a data set containing collections and their constituent sets
(or collections) created by a subset of Flickr users who are members
of seventeen public groups devoted to wildlife and nature
photography~\cite{www09folksonomies}. These users had many other
common interests,
such as travel and sports, arts and crafts, and
people and portraiture.
We extracted all the tags associated with images in the set,
and retrieved all other images that the user annotated with these tags.
%
We constructed personal hierarchies, or saplings, from this data, with each sapling rooted at one of user's top-level collections. For reasons described in \secref{sec:preproc}, we ignore collections with composite names. This reduces the size of the data set to $20,759$ saplings created by $7,121$ users.
A small number of these saplings are multi-level.


\begin{table}

\begin{tabular}{|llll}
\hline
\multicolumn{1}{|l|}{\scriptsize{Parameters}} & \multicolumn{3}{c|}{\scriptsize{Description}} \\ 
\hline
\multicolumn{1}{|l|}{\scriptsize{\texttt{K}}} & \multicolumn{3}{l|}{\scriptsize{The number of top frequent tags}} \\ 
\hline
\multicolumn{1}{|l|}{\scriptsize{\texttt{J}}} & \multicolumn{3}{p{2.4in}|}{\scriptsize{The number of common tags for tag similarity}} \\ 
\hline
\multicolumn{1}{|l|}{\scriptsize{$\alpha_{RR}$}} & \multicolumn{3}{p{2.4in}|}{\scriptsize{The weight combination of local and structural similarity for computing root-to-root similarity}} \\ 
\hline
\multicolumn{1}{|l|}{\scriptsize{$\alpha_{LR}$}} & \multicolumn{3}{p{2.4in}|}{\scriptsize{The weight combination of local and structural similarity for computing leaf-to-root similarity}} \\ 
\hline
\multicolumn{1}{|l|}{\scriptsize{$\beta$}} & \multicolumn{3}{p{2.4in}|}{\scriptsize{The weight combination of name and tag similarity (not required in our experiment)}} \\ 
\hline
\multicolumn{1}{|l|}{\scriptsize{$\tau$}} & \multicolumn{3}{p{2.4in}|}{\scriptsize{The similarity threshold}} \\ 
\hline
\end{tabular}
\caption{Parameters of the folksonomy learning approach.}
\label{tbl:param}
\end{table}

The folksonomy learning approach described in this paper has
a number of parameters as shown in~\tabref{tbl:param}. In our experiment, we ignored the parameter $\beta$ since only sapling nodes with the same name are needed to be compared as described the previous section. 
To explore the range of these parameters, we set up a small experiment by first selecting $5$ different seed terms\footnote{\term{ski}, \term{bird}, \term{victoria}, \term{africa} and \term{insect}}; then running the approach with different values.\comment{
of these parameters to induce folksonomies from the seeds.} Optimal  parameter values would enable the approach to reasonably combine/separate saplings with similar/different senses. We manually inspected the induced folksonomies to check how the saplings were merged/separated.

The parameter $\texttt{K}$  allows the approach to consider only top frequency tags, which tend to be more stable and less noisy\comment{than the less frequent ones}~\cite{Golder06}. Nevertheless, the top tags will not contain enough information if the number is set too low, e.g., $K=10$. At the fixed values of the common tag threshold, $\texttt{J} = 4$, and the structural-local weight combination, $\alpha_{RR} = 0.1$ (in this case, we simply evaluated on merging root-root nodes; hence there is no need for $\alpha_{LR}$), we found that the approach performs reasonably well when the value of $\texttt{K}$ is around $30$--$60$, while the performance starts to degrade for $\texttt{K} > 60$. Smaller values of $\texttt{J}$ leads to a weak  tag similarity measure, which, in turn, mistakenly causes the approach to merge saplings with different senses. Large $\texttt{J}$ will be relatively stringent, and as a result, saplings of the same sense will not be merged. We found that, at $\texttt{K} = 40$, the value of $\texttt{J}$ between $4$ to $6$ allows reasonable results. 

For $\alpha_{RR}$ and $\alpha_{LR}$, the weight combination between local and structural similarity for root-root and leaf-root nodes in \eqrefx{eq:grandfunction}, the larger the values the more the similarity measure emphasizes on the structural similarity. From our experiments, we found that the structure information is very informative. When $\alpha_{RR}$ is set to a very large value or the maximum, $1.0$, the approach clusters ``structure-rich'' saplings, i.e., saplings containing many children, reasonably well. For leaf-to-root merging or in situations where structural information is uncommon, local similarity becomes more important. We discovered that at $\alpha_{RR} = 0.1$ and $\alpha_{LR}=0.8$, the approach produces reasonable folksonomies. Due to space limitations, we do not include the complete set of results. Here, we report the parameter values that resulted in good performance: we set $\texttt{K}=40$; $\texttt{J}=4$\comment{; $\alpha_{RR}=0.9$; $\alpha_{LR}=0.2$}. In addition, since all similarity measures are normalized to range within $0.0$ and $1.0$, we set $\tau=0.5$. 



\comment{
We explored a range of parameter values; due to space limitations, we do not include the complete set of
results. Here, we report results obtained through a combination that resulted in good performance: we set $\alpha = 0.4$ the weight of the structural similarity for root-to-root, and $0.6$ for root-to-leaf similarity, we use
$0.5$ as the similarity threshold for deciding whether to merge nodes,
the number of top tags retained in each
node is $\texttt{K}=20$, the number of common tags used to compute tag
similarity is $\texttt{J}=0.10\texttt{K}$, and the number of tags used
in
continuity checking was $\texttt{L}=0.05\texttt{K}$.
The rationale for selecting the different values for $\alpha$ is that,
the merging of roots is based more on overlap between child nodes;
hence, we can rely more on the structural similarity. In the case of
root-to-leaf, we instead set more weight on local similarity of common
tags.
}

We compare \algname{sap} against the folksonomy learning method, \algname{sig}, described in \cite{www09folksonomies}. Briefly, \algname{sig} 
first breaks a given sapling into
(collection-set) individual parent-child relations. With the assumption that the nodes with
the same (stemmed) name refer to the same concept, the approach
employs hypothesis testing to identify the informative
relations, i.e., checking if the relation is not generated at random. Informative relations are then linked into a deeper
folksonomy.
We used a significance test threshold of $0.01$.
\subsection{Methodology}
We quantitatively evaluate the induced folksonomies by (1) automatically comparing
them to a reference hierarchy;
(2) structural evaluation; (3) manual evaluation.

\noindent \textbf{Evaluation against the reference hierarchy:} We use the reference hierarchy from the Open Directory
Project\\(ODP).\footnote{http://rdf.dmoz.org/, as of September 2008}
We selected ODP because, in contrast to
WordNet, ODP is generated, reviewed and revised by many registered
users. These users seem to use more colloquial terms than appear in
WordNet. In addition, like Flickr users, they specify less formal
relations, mainly broader/narrower relations. WordNet, on the other
hand, specifies a number of formal relations among concepts, including
hypernymy and meronymy.

\comment{
For the hand labeled data, we use 3 human
subjects to evaluate the portions of induced folksonomies, which were
not comparable to ODP hierarchy.}
We use methodology described in \cite{www09folksonomies} to
automatically evaluate the quality of the learned
folksonomies. Although ODP and saplings are generated from different
sources, there is substantial vocabulary overlap that makes them
comparable.  Since the ODP hierarchy is relatively large and composed
of many topics, we had to carve out the ``relevant'' portion for
comparison. First, we specified a seed, ${S}$, which is the root of
the learned folksonomy $\mathds{F}$ and the reference hierarchy to
which it is compared.

Next, the folksonomy is expanded two levels along the relations in
$\mathds{F}$. The nodes in the second level are added as
leaf candidates, $LC$. If the spanning stops after one level, we also
add this node's name to $LC$.
Given $S$ and $LC$, we identify leaf
candidates, $LCD$, that also appear in ODP, $\mathds{D}$. All paths
from $S$ to $LCD$ in $\mathds{D}$ constitute the reference
hierarchy for the seed ${S}$.

%
Next, ${S}$ is used as seed for learning the folksonomy associated with this
concept. In \algname{sig}, ${S}$ and $LC$ are both used to learn the
folksonomy.  The maximum depth of learned trees is limited to $4$. The
metrics to compare the learned folksonomies to the reference are
\emph{Lexical Recall}~\cite{MaedcheS02} and the modified
\emph{Taxonomic Overlap} defined in \cite{www09folksonomies}, $mTO$.
Lexical Recall measures the overlap
between the learned and reference taxonomies, independent of their
structure. $mTO$ measures the quality of structural
alignment of the taxonomies.
Here, we report the harmonic mean, $fmTO$, instead, because of $mTO$'s asymmetry.
Since the proposed approach generates bushy folksonomies
whose leaf nodes may not appear in the reference taxonomy, the $mTO$ metric may unfairly penalize the learned folksonomy. Instead, we only consider the paths of the learned folksonomy that are comparable to the reference hierarchy. Specifically, for each leaf $l$ in $LCD$, we select the path $S \rightarrow l$ in the learned folksonomy and compare it to one in the reference hierarchy. If there are many comparable paths existing in the reference, we select the one that has the highest $LR$ to compare.
%

\comment{In addition to $LR$ and $fmTO$, which measure how an induced tree consistent with the gold-standard hierarchy, a}
\noindent \textbf{Structural evaluation:} Ideally, we prefer an approach that generates bushier and deeper trees. The scope of concepts in such trees are broadly enumerated (tree width); while, each concept is subcategorized in enough detail (tree depth). Although one can use an average depth  of a tree and branching factor, it is difficult to justify which trees are better overall since these metrics are independent. A very bushy tree may have only 1 level depth; meanwhile, a very deep tree may have a chain-like structure. In this work, we define a simple, yet intuitive measure, Area Under Tree (AUT), which takes \emph{both} tree bushiness and depth into account. To calculate AUT for a certain tree, we compute the distribution of the number of nodes in each level and then compute the area under the distribution. Intuitively, trees that keep branching out at each level will have larger AUT than those that are short and thin. Suppose that we have a tree with one node at the root, three nodes at $1^{st}$ level and four at $2^{nd}$. With the scale of tree depth set to $1.0$, AUT of this tree would be $0.5\times(1+3) + 0.5\times(3+4) = 5.5$ (a sum of trapezoids).  

\noindent \textbf{Manual evaluation:} We use 3 human
subjects to evaluate the portions of induced folksonomies which were
not comparable to ODP hierarchy. We randomly selected $10\%$ of the paths (all of them if there are
fewer than $10$ paths in the learned folksonomy) that are not in the
reference hierarchy and asked three judges to evaluate them. If a
portion of the path is incorrect, either because an incorrect concept
appears or the ordering of concepts is wrong, the judges were
asked to mark it incorrect, otherwise it is correct. They can also
mark the path ``unsure'' if there is not enough evidence for a decision.
A path's label is based on the majority decision. If there is no
agreement, or the path is marked uncertain by all judges, we
exclude it.


\begin{table*}[tbh]
\centering
\setlength{\tabcolsep}{1pt}
\scriptsize{
\begin{tabular}{|lllllllllllllll}
\hline
\multicolumn{1}{|l|}{} & \multicolumn{4}{c|}{Whole folksonomies} & \multicolumn{8}{c|}{Comparison with ODP} & \multicolumn{2}{c|}{Manual} \\ 
\cline{2-15}
\multicolumn{1}{|l|}{} & \multicolumn{2}{c|}{\#leaves} & \multicolumn{2}{c|}{AUT} & \multicolumn{2}{c|}{\#ovlp lvs} & \multicolumn{2}{c|}{fmTO} & \multicolumn{2}{c|}{LR} & \multicolumn{2}{c|}{AUT} & \multicolumn{2}{c|}{Acc (10\%)} \\ 
\cline{2-15}
\multicolumn{1}{|c|}{seeds} & \multicolumn{1}{c|}{sig} & \multicolumn{1}{c|}{sap} & \multicolumn{1}{c|}{sig} & \multicolumn{1}{c|}{sap} & \multicolumn{1}{c|}{sig} & \multicolumn{1}{c|}{sap} & \multicolumn{1}{c|}{sig} & \multicolumn{1}{c|}{sap} & \multicolumn{1}{c|}{sig} & \multicolumn{1}{c|}{sap} & \multicolumn{1}{c|}{sig} & \multicolumn{1}{c|}{sap} & \multicolumn{1}{c|}{sig} & \multicolumn{1}{c|}{sap} \\ 
\hline
\multicolumn{1}{|l|}{anim} & \multicolumn{1}{r|}{268} & \multicolumn{1}{r|}{\textbf{583}} & \multicolumn{1}{r|}{694.0} & \multicolumn{1}{r|}{\textbf{1076.0}} & \multicolumn{1}{r|}{68} & \multicolumn{1}{r|}{\textbf{92}} & \multicolumn{1}{r|}{0.602} & \multicolumn{1}{r|}{\textbf{0.659}} & \multicolumn{1}{r|}{0.281} & \multicolumn{1}{r|}{\textbf{0.360}} & \multicolumn{1}{r|}{160.0} & \multicolumn{1}{r|}{\textbf{189.5}} & \multicolumn{1}{r|}{\textbf{0.89}} & \multicolumn{1}{r|}{0.74} \\
\hline
\multicolumn{1}{|l|}{bird} & \multicolumn{1}{r|}{73} & \multicolumn{1}{r|}{\textbf{103}} & \multicolumn{1}{r|}{84.5} & \multicolumn{1}{r|}{\textbf{113.5}} & \multicolumn{1}{r|}{20} & \multicolumn{1}{r|}{\textbf{22}} & \multicolumn{1}{r|}{\textbf{0.760}} & \multicolumn{1}{r|}{0.755} & \multicolumn{1}{r|}{0.281} & \multicolumn{1}{r|}{\textbf{0.315}} & \multicolumn{1}{r|}{21.5} & \multicolumn{1}{r|}{\textbf{28.5}} & \multicolumn{1}{r|}{0.60} & \multicolumn{1}{r|}{\textbf{1.00}} \\
\hline
\multicolumn{1}{|l|}{invertebr} & \multicolumn{1}{r|}{11} & \multicolumn{1}{r|}{\textbf{15}} & \multicolumn{1}{r|}{15.5} & \multicolumn{1}{r|}{\textbf{19.5}} & \multicolumn{1}{r|}{\textbf{3}} & \multicolumn{1}{r|}{1} & \multicolumn{1}{r|}{0.762} & \multicolumn{1}{r|}{\textbf{1.000}} & \multicolumn{1}{r|}{\textbf{0.250}} & \multicolumn{1}{r|}{0.125} & \multicolumn{1}{r|}{\textbf{4.5}} & \multicolumn{1}{r|}{1.5} & \multicolumn{1}{r|}{1.00} & \multicolumn{1}{r|}{1.00} \\
\hline
\multicolumn{1}{|l|}{vertebr} & \multicolumn{1}{r|}{80} & \multicolumn{1}{r|}{\textbf{114}} & \multicolumn{1}{r|}{162.5} & \multicolumn{1}{r|}{\textbf{236.5}} & \multicolumn{1}{r|}{\textbf{1}} & \multicolumn{1}{r|}{0} & \multicolumn{1}{r|}{1.000} & \multicolumn{1}{r|}{n/a} & \multicolumn{1}{r|}{\textbf{0.600}} & \multicolumn{1}{r|}{0.200} & \multicolumn{1}{r|}{2.5} & \multicolumn{1}{r|}{n/a} & \multicolumn{1}{r|}{1.00} & \multicolumn{1}{r|}{1.00} \\
\hline
\multicolumn{1}{|l|}{insect} & \multicolumn{1}{r|}{29} & \multicolumn{1}{r|}{\textbf{44}} & \multicolumn{1}{r|}{35.5} & \multicolumn{1}{r|}{\textbf{61.5}} & \multicolumn{1}{r|}{5} & \multicolumn{1}{r|}{5} & \multicolumn{1}{r|}{0.924} & \multicolumn{1}{r|}{0.924} & \multicolumn{1}{r|}{0.857} & \multicolumn{1}{r|}{0.857} & \multicolumn{1}{r|}{6.5} & \multicolumn{1}{r|}{6.5} & \multicolumn{1}{r|}{1.00} & \multicolumn{1}{r|}{1.00} \\
\hline
\multicolumn{1}{|l|}{fish} & \multicolumn{1}{r|}{\textbf{7}} & \multicolumn{1}{r|}{6} & \multicolumn{1}{r|}{\textbf{7.5}} & \multicolumn{1}{r|}{6.5} & \multicolumn{1}{r|}{0} & \multicolumn{1}{r|}{0} & \multicolumn{1}{r|}{n/a} & \multicolumn{1}{r|}{n/a} & \multicolumn{1}{r|}{0.016} & \multicolumn{1}{r|}{0.016} & \multicolumn{1}{r|}{n/a} & \multicolumn{1}{r|}{n/a} & \multicolumn{1}{r|}{1.00} & \multicolumn{1}{r|}{1.00} \\
\hline
\multicolumn{1}{|l|}{plant} & \multicolumn{1}{r|}{110} & \multicolumn{1}{r|}{\textbf{194}} & \multicolumn{1}{r|}{265.5} & \multicolumn{1}{r|}{\textbf{426.0}} & \multicolumn{1}{r|}{6} & \multicolumn{1}{r|}{\textbf{7}} & \multicolumn{1}{r|}{0.613} & \multicolumn{1}{r|}{\textbf{0.735}} & \multicolumn{1}{r|}{0.250} & \multicolumn{1}{r|}{\textbf{0.273}} & \multicolumn{1}{r|}{\textbf{13.0}} & \multicolumn{1}{r|}{11.5} & \multicolumn{1}{r|}{0.67} & \multicolumn{1}{r|}{\textbf{1.00}} \\
\hline
\multicolumn{1}{|l|}{flora} & \multicolumn{1}{r|}{64} & \multicolumn{1}{r|}{\textbf{403}} & \multicolumn{1}{r|}{173.0} & \multicolumn{1}{r|}{\textbf{1048.5}} & \multicolumn{1}{r|}{6} & \multicolumn{1}{r|}{\textbf{18}} & \multicolumn{1}{r|}{\textbf{0.483}} & \multicolumn{1}{r|}{0.481} & \multicolumn{1}{r|}{0.130} & \multicolumn{1}{r|}{\textbf{0.407}} & \multicolumn{1}{r|}{16.0} & \multicolumn{1}{r|}{\textbf{84.0}} & \multicolumn{1}{r|}{1.00} & \multicolumn{1}{r|}{1.00} \\
\hline
\multicolumn{1}{|l|}{fauna} & \multicolumn{1}{r|}{141} & \multicolumn{1}{r|}{\textbf{609}} & \multicolumn{1}{r|}{420.0} & \multicolumn{1}{r|}{\textbf{1146.0}} & \multicolumn{1}{r|}{9} & \multicolumn{1}{r|}{\textbf{31}} & \multicolumn{1}{r|}{0.463} & \multicolumn{1}{r|}{\textbf{0.490}} & \multicolumn{1}{r|}{0.113} & \multicolumn{1}{r|}{\textbf{0.212}} & \multicolumn{1}{r|}{27.0} & \multicolumn{1}{r|}{\textbf{71.5}} & \multicolumn{1}{r|}{\textbf{0.91}} & \multicolumn{1}{r|}{0.85} \\
\hline
\multicolumn{1}{|l|}{flower} & \multicolumn{1}{r|}{112} & \multicolumn{1}{r|}{\textbf{169}} & \multicolumn{1}{r|}{210.5} & \multicolumn{1}{r|}{\textbf{226.5}} & \multicolumn{1}{r|}{1} & \multicolumn{1}{r|}{1} & \multicolumn{1}{r|}{0.379} & \multicolumn{1}{r|}{\textbf{1.000}} & \multicolumn{1}{r|}{\textbf{0.267}} & \multicolumn{1}{r|}{0.250} & \multicolumn{1}{r|}{\textbf{3.5}} & \multicolumn{1}{r|}{1.5} & \multicolumn{1}{r|}{1.00} & \multicolumn{1}{r|}{n/a} \\
\hline
\multicolumn{1}{|l|}{reptil} & \multicolumn{1}{r|}{3} & \multicolumn{1}{r|}{\textbf{4}} & \multicolumn{1}{r|}{4.5} & \multicolumn{1}{r|}{4.5} & \multicolumn{1}{r|}{2} & \multicolumn{1}{r|}{\textbf{3}} & \multicolumn{1}{r|}{\textbf{0.625}} & \multicolumn{1}{r|}{0.622} & \multicolumn{1}{r|}{0.500} & \multicolumn{1}{r|}{\textbf{0.667}} & \multicolumn{1}{r|}{2.5} & \multicolumn{1}{r|}{\textbf{3.5}} & \multicolumn{1}{r|}{n/a} & \multicolumn{1}{r|}{n/a} \\
\hline
\multicolumn{1}{|l|}{amphibian} & \multicolumn{1}{r|}{1} & \multicolumn{1}{r|}{1} & \multicolumn{1}{r|}{1.5} & \multicolumn{1}{r|}{1.5} & \multicolumn{1}{r|}{1} & \multicolumn{1}{r|}{1} & \multicolumn{1}{r|}{1.000} & \multicolumn{1}{r|}{1.000} & \multicolumn{1}{r|}{1.000} & \multicolumn{1}{r|}{1.000} & \multicolumn{1}{r|}{1.5} & \multicolumn{1}{r|}{1.5} & \multicolumn{1}{r|}{n/a} & \multicolumn{1}{r|}{n/a} \\
\hline
\multicolumn{1}{|l|}{build} & \multicolumn{1}{r|}{7} & \multicolumn{1}{r|}{\textbf{23}} & \multicolumn{1}{r|}{11.5} & \multicolumn{1}{r|}{\textbf{37.5}} & \multicolumn{1}{r|}{0} & \multicolumn{1}{r|}{0} & \multicolumn{1}{r|}{n/a} & \multicolumn{1}{r|}{n/a} & \multicolumn{1}{r|}{1.000} & \multicolumn{1}{r|}{1.000} & \multicolumn{1}{r|}{n/a} & \multicolumn{1}{r|}{n/a} & \multicolumn{1}{r|}{1.00} & \multicolumn{1}{r|}{1.00} \\
\hline
\multicolumn{1}{|l|}{urban} & \multicolumn{1}{r|}{6} & \multicolumn{1}{r|}{\textbf{80}} & \multicolumn{1}{r|}{15.0} & \multicolumn{1}{r|}{\textbf{145.5}} & \multicolumn{1}{r|}{0} & \multicolumn{1}{r|}{0} & \multicolumn{1}{r|}{n/a} & \multicolumn{1}{r|}{n/a} & \multicolumn{1}{r|}{0.071} & \multicolumn{1}{r|}{0.071} & \multicolumn{1}{r|}{n/a} & \multicolumn{1}{r|}{n/a} & \multicolumn{1}{r|}{1.00} & \multicolumn{1}{r|}{1.00} \\
\hline
\multicolumn{1}{|l|}{countri} & \multicolumn{1}{r|}{378} & \multicolumn{1}{r|}{\textbf{1605}} & \multicolumn{1}{r|}{798.5} & \multicolumn{1}{r|}{\textbf{4504.0}} & \multicolumn{1}{r|}{2} & \multicolumn{1}{r|}{\textbf{4}} & \multicolumn{1}{r|}{0.447} & \multicolumn{1}{r|}{\textbf{0.665}} & \multicolumn{1}{r|}{0.143} & \multicolumn{1}{r|}{\textbf{0.214}} & \multicolumn{1}{r|}{8.0} & \multicolumn{1}{r|}{\textbf{8.5}} & \multicolumn{1}{r|}{1.00} & \multicolumn{1}{r|}{1.00} \\
\hline
\multicolumn{1}{|l|}{africa} & \multicolumn{1}{r|}{53} & \multicolumn{1}{r|}{\textbf{71}} & \multicolumn{1}{r|}{90.5} & \multicolumn{1}{r|}{\textbf{119.5}} & \multicolumn{1}{r|}{23} & \multicolumn{1}{r|}{\textbf{27}} & \multicolumn{1}{r|}{0.773} & \multicolumn{1}{r|}{\textbf{0.895}} & \multicolumn{1}{r|}{0.508} & \multicolumn{1}{r|}{\textbf{0.547}} & \multicolumn{1}{r|}{37.5} & \multicolumn{1}{r|}{\textbf{40.5}} & \multicolumn{1}{r|}{1.00} & \multicolumn{1}{r|}{1.00} \\
\hline
\multicolumn{1}{|l|}{asia} & \multicolumn{1}{r|}{187} & \multicolumn{1}{r|}{\textbf{284}} & \multicolumn{1}{r|}{389.0} & \multicolumn{1}{r|}{\textbf{631.5}} & \multicolumn{1}{r|}{80} & \multicolumn{1}{r|}{\textbf{85}} & \multicolumn{1}{r|}{0.734} & \multicolumn{1}{r|}{\textbf{0.788}} & \multicolumn{1}{r|}{0.396} & \multicolumn{1}{r|}{\textbf{0.484}} & \multicolumn{1}{r|}{165.5} & \multicolumn{1}{r|}{\textbf{168.5}} & \multicolumn{1}{r|}{1.00} & \multicolumn{1}{r|}{1.00} \\
\hline
\multicolumn{1}{|l|}{europ} & \multicolumn{1}{r|}{379} & \multicolumn{1}{r|}{\textbf{1073}} & \multicolumn{1}{r|}{916.0} & \multicolumn{1}{r|}{\textbf{2706.5}} & \multicolumn{1}{r|}{165} & \multicolumn{1}{r|}{\textbf{301}} & \multicolumn{1}{r|}{0.619} & \multicolumn{1}{r|}{\textbf{0.670}} & \multicolumn{1}{r|}{0.236} & \multicolumn{1}{r|}{\textbf{0.418}} & \multicolumn{1}{r|}{369.0} & \multicolumn{1}{r|}{\textbf{874.5}} & \multicolumn{1}{r|}{\textbf{1.00}} & \multicolumn{1}{r|}{0.94} \\
\hline
\multicolumn{1}{|l|}{south africa} & \multicolumn{1}{r|}{12} & \multicolumn{1}{r|}{\textbf{17}} & \multicolumn{1}{r|}{15.5} & \multicolumn{1}{r|}{\textbf{18.5}} & \multicolumn{1}{r|}{3} & \multicolumn{1}{r|}{3} & \multicolumn{1}{r|}{0.431} & \multicolumn{1}{r|}{\textbf{0.600}} & \multicolumn{1}{r|}{0.444} & \multicolumn{1}{r|}{0.444} & \multicolumn{1}{r|}{3.5} & \multicolumn{1}{r|}{3.5} & \multicolumn{1}{r|}{0.78} & \multicolumn{1}{r|}{\textbf{1.00}} \\
\hline
\multicolumn{1}{|l|}{north america} & \multicolumn{1}{r|}{166} & \multicolumn{1}{r|}{\textbf{731}} & \multicolumn{1}{r|}{435.0} & \multicolumn{1}{r|}{\textbf{2203.5}} & \multicolumn{1}{r|}{67} & \multicolumn{1}{r|}{\textbf{118}} & \multicolumn{1}{r|}{0.545} & \multicolumn{1}{r|}{\textbf{0.576}} & \multicolumn{1}{r|}{0.165} & \multicolumn{1}{r|}{\textbf{0.319}} & \multicolumn{1}{r|}{170.5} & \multicolumn{1}{r|}{\textbf{361.5}} & \multicolumn{1}{r|}{\textbf{1.00}} & \multicolumn{1}{r|}{0.92} \\
\hline
\multicolumn{1}{|l|}{south america} & \multicolumn{1}{r|}{32} & \multicolumn{1}{r|}{\textbf{50}} & \multicolumn{1}{r|}{54.5} & \multicolumn{1}{r|}{\textbf{101.5}} & \multicolumn{1}{r|}{12} & \multicolumn{1}{r|}{\textbf{15}} & \multicolumn{1}{r|}{0.706} & \multicolumn{1}{r|}{\textbf{0.832}} & \multicolumn{1}{r|}{0.415} & \multicolumn{1}{r|}{\textbf{0.463}} & \multicolumn{1}{r|}{20.5} & \multicolumn{1}{r|}{\textbf{28.5}} & \multicolumn{1}{r|}{1.00} & \multicolumn{1}{r|}{1.00} \\
\hline
\multicolumn{1}{|l|}{central america} & \multicolumn{1}{r|}{\textbf{27}} & \multicolumn{1}{r|}{8} & \multicolumn{1}{r|}{\textbf{53.5}} & \multicolumn{1}{r|}{12.5} & \multicolumn{1}{r|}{1} & \multicolumn{1}{r|}{\textbf{2}} & \multicolumn{1}{r|}{0.631} & \multicolumn{1}{r|}{\textbf{0.754}} & \multicolumn{1}{r|}{0.417} & \multicolumn{1}{r|}{\textbf{0.500}} & \multicolumn{1}{r|}{2.5} & \multicolumn{1}{r|}{\textbf{4.5}} & \multicolumn{1}{r|}{1.00} & \multicolumn{1}{r|}{1.00} \\
\hline
\multicolumn{1}{|l|}{unit kingdom} & \multicolumn{1}{r|}{106} & \multicolumn{1}{r|}{\textbf{267}} & \multicolumn{1}{r|}{274.5} & \multicolumn{1}{r|}{\textbf{658.5}} & \multicolumn{1}{r|}{31} & \multicolumn{1}{r|}{\textbf{82}} & \multicolumn{1}{r|}{\textbf{0.787}} & \multicolumn{1}{r|}{0.724} & \multicolumn{1}{r|}{0.099} & \multicolumn{1}{r|}{\textbf{0.127}} & \multicolumn{1}{r|}{71.5} & \multicolumn{1}{r|}{\textbf{179.5}} & \multicolumn{1}{r|}{1.00} & \multicolumn{1}{r|}{1.00} \\
\hline
\multicolumn{1}{|l|}{unit state} & \multicolumn{1}{r|}{102} & \multicolumn{1}{r|}{\textbf{375}} & \multicolumn{1}{r|}{217.0} & \multicolumn{1}{r|}{\textbf{936.5}} & \multicolumn{1}{r|}{35} & \multicolumn{1}{r|}{55} & \multicolumn{1}{r|}{0.620} & \multicolumn{1}{r|}{\textbf{0.749}} & \multicolumn{1}{r|}{0.130} & \multicolumn{1}{r|}{\textbf{0.256}} & \multicolumn{1}{r|}{74.5} & \multicolumn{1}{r|}{\textbf{122.0}} & \multicolumn{1}{r|}{1.00} & \multicolumn{1}{r|}{1.00} \\
\hline
\multicolumn{1}{|l|}{world} & \multicolumn{1}{r|}{545} & \multicolumn{1}{r|}{\textbf{3177}} & \multicolumn{1}{r|}{1437.0} & \multicolumn{1}{r|}{\textbf{9235.0}} & \multicolumn{1}{r|}{191} & \multicolumn{1}{r|}{\textbf{475}} & \multicolumn{1}{r|}{\textbf{0.476}} & \multicolumn{1}{r|}{0.461} & \multicolumn{1}{r|}{0.085} & \multicolumn{1}{r|}{\textbf{0.215}} & \multicolumn{1}{r|}{490.0} & \multicolumn{1}{r|}{\textbf{1676.5}} & \multicolumn{1}{r|}{\textbf{0.97}} & \multicolumn{1}{r|}{0.96} \\
\hline
\multicolumn{1}{|l|}{citi} & \multicolumn{1}{r|}{123} & \multicolumn{1}{r|}{\textbf{448}} & \multicolumn{1}{r|}{234.0} & \multicolumn{1}{r|}{\textbf{927.5}} & \multicolumn{1}{r|}{0} & \multicolumn{1}{r|}{0} & \multicolumn{1}{r|}{n/a} & \multicolumn{1}{r|}{n/a} & \multicolumn{1}{r|}{\textbf{0.111}} & \multicolumn{1}{r|}{0.100} & \multicolumn{1}{r|}{n/a} & \multicolumn{1}{r|}{2.5} & \multicolumn{1}{r|}{1.00} & \multicolumn{1}{r|}{1.00} \\
\hline
\multicolumn{1}{|l|}{craft} & \multicolumn{1}{r|}{\textbf{5}} & \multicolumn{1}{r|}{1} & \multicolumn{1}{r|}{\textbf{10.5}} & \multicolumn{1}{r|}{1.5} & \multicolumn{1}{r|}{1} & \multicolumn{1}{r|}{0} & \multicolumn{1}{r|}{0.603} & \multicolumn{1}{r|}{n/a} & \multicolumn{1}{r|}{\textbf{0.056}} & \multicolumn{1}{r|}{0.050} & \multicolumn{1}{r|}{2.5} & \multicolumn{1}{r|}{n/a} & \multicolumn{1}{r|}{1.00} & \multicolumn{1}{r|}{n/a} \\
\hline
\multicolumn{1}{|l|}{dog} & \multicolumn{1}{r|}{15} & \multicolumn{1}{r|}{\textbf{26}} & \multicolumn{1}{r|}{17.5} & \multicolumn{1}{r|}{\textbf{28.5}} & \multicolumn{1}{r|}{0} & \multicolumn{1}{r|}{\textbf{1}} & \multicolumn{1}{r|}{n/a} & \multicolumn{1}{r|}{1.000} & \multicolumn{1}{r|}{0.045} & \multicolumn{1}{r|}{\textbf{0.080}} & \multicolumn{1}{r|}{n/a} & \multicolumn{1}{r|}{1.5} & \multicolumn{1}{r|}{n/a} & \multicolumn{1}{r|}{n/a} \\
\hline
\multicolumn{1}{|l|}{cat} & \multicolumn{1}{r|}{11} & \multicolumn{1}{r|}{\textbf{39}} & \multicolumn{1}{r|}{13.5} & \multicolumn{1}{r|}{\textbf{41.5}} & \multicolumn{1}{r|}{0} & \multicolumn{1}{r|}{0} & \multicolumn{1}{r|}{n/a} & \multicolumn{1}{r|}{n/a} & \multicolumn{1}{r|}{0.100} & \multicolumn{1}{r|}{0.100} & \multicolumn{1}{r|}{n/a} & \multicolumn{1}{r|}{n/a} & \multicolumn{1}{r|}{n/a} & \multicolumn{1}{r|}{n/a} \\
\hline
\multicolumn{1}{|l|}{sport} & \multicolumn{1}{r|}{\textbf{207}} & \multicolumn{1}{r|}{74} & \multicolumn{1}{r|}{\textbf{407.0}} & \multicolumn{1}{r|}{86.5} & \multicolumn{1}{r|}{19} & \multicolumn{1}{r|}{\textbf{27}} & \multicolumn{1}{r|}{\textbf{0.693}} & \multicolumn{1}{r|}{0.647} & \multicolumn{1}{r|}{\textbf{0.091}} & \multicolumn{1}{r|}{0.084} & \multicolumn{1}{r|}{30.0} & \multicolumn{1}{r|}{\textbf{31.5}} & \multicolumn{1}{r|}{0.28} & \multicolumn{1}{r|}{\textbf{1.00}} \\
\hline
\multicolumn{1}{|l|}{australia} & \multicolumn{1}{r|}{47} & \multicolumn{1}{r|}{\textbf{83}} & \multicolumn{1}{r|}{71.0} & \multicolumn{1}{r|}{\textbf{147.5}} & \multicolumn{1}{r|}{12} & \multicolumn{1}{r|}{\textbf{27}} & \multicolumn{1}{r|}{0.354} & \multicolumn{1}{r|}{\textbf{0.665}} & \multicolumn{1}{r|}{0.123} & \multicolumn{1}{r|}{\textbf{0.216}} & \multicolumn{1}{r|}{14.5} & \multicolumn{1}{r|}{\textbf{36.5}} & \multicolumn{1}{r|}{0.67} & \multicolumn{1}{r|}{\textbf{1.00}} \\
\hline
\multicolumn{1}{|l|}{canada} & \multicolumn{1}{r|}{55} & \multicolumn{1}{r|}{\textbf{763}} & \multicolumn{1}{r|}{128.0} & \multicolumn{1}{r|}{\textbf{2502.0}} & \multicolumn{1}{r|}{11} & \multicolumn{1}{r|}{\textbf{27}} & \multicolumn{1}{r|}{\textbf{0.620}} & \multicolumn{1}{r|}{0.587} & \multicolumn{1}{r|}{0.158} & \multicolumn{1}{r|}{\textbf{0.241}} & \multicolumn{1}{r|}{21.5} & \multicolumn{1}{r|}{\textbf{75.5}} & \multicolumn{1}{r|}{1.00} & \multicolumn{1}{r|}{1.00} \\

\hline
\end{tabular}
}

\caption {This table presents empirical validation on folksonomies induced by the proposed approach, \algname{sap}, comparing to the baseline approach, \algname{sig}. The first column group presents properties of the whole induced trees: the number of leaves and Area Under Tree(AUT). The second column group reports the quality of induced trees, relatively to the ODP hierarchy. The metrics in this group are \emph{modified Taxonomic Overlap} (\emph{fmTO}) (averaged using Harmonic Mean), \emph{Lexical Recall} (\emph{LR}), where their scales are ranging from $0.0$ to $1.0$ (the more the better), as AUT is computed from portions of the trees, which are comparable to ODP. ``\#ovlp lvs'' stands for a number of overlap leaves (to ODP).  The last column group reports performance on manually labeled portions of the trees, which do not occur in ODP.  In some cases, ``n/a'' exists since we cannot compute its corresponding value.}
\label{tbl:results}
\end{table*}

\subsection{Results} In \tabref{tbl:results}, we compare the quality of the
folksonomy learned for each seed by \algname{sap},
and the earlier work,
\algname{sig}. \algname{sap} generally recovers a larger number of concepts, relative to ODP, as indicated by the numbers of overlapping leaves (in $90\%$ of the cases) and better LR scores (in $76\%$ of the cases). Moreover, \algname{sap} can produce trees with higher quality, relative to the ODP, as indicated by $fmTO$ score (in $68\%$ of the cases). From the structural evaluation, \algname{sap} produced bushier trees as indicated by AUT in $87\%$ of the cases. In addition, the average depth (not shown in the Table) from roots to all leaves of the trees over all cases generated by \algname{sap} is deeper than \algname{sig} ($2.68$ vs. $2.37$).  

\comment{
\algname{sap} produces bushier trees because individual saplings will be judged relevant using structural information, rather than frequencies of individual relations as in \algname{sig}. Although \algname{sig} can remove many idiosyncratic relations, it also removes many of informative ones too.
}

Although the manual evaluation suggests that both approaches can induce about the same quality on the paths that are uncomparable to ODP, after closely inspecting the learned trees, we found that \algname{sap} demonstrates its advantage over \algname{sig} in
disambiguating and correctly attaching relevant
saplings to appropriate induced trees. For instance, \term{bird} tree
produced by
\algname{sap} does
not includes \term{Istanbul} or other Turkey locations, as shown in \figref{fig:trees}. 
%
%
In the \term{sport} tree, \algname{sap} does not include any concept about the \term{sky} (Note that skies and skiing share common  name). In addition, there are no concepts about irrelevant
events like birthdays and parades appearing in the
tree.
There are some cases, e.g., \term{dog} and \term{cat}, where
we could not compute the hand labeling scores because
these trees often contained pet names, rather than breeds.


We further considered how many of the
incorrect paths
are caused by node
ambiguity. To do so, we first identified ambiguous terms,
and checked to see how many of the incorrect paths contain these
terms. Although it is not obvious how to automatically identify ambiguous
terms, we use the following heuristic to determine the possible
ambiguities: for a given leaf of the induced tree,
if many different merged senses exist
(i.e., $>10$), then we consider the leaf ambiguous.  During the tree
induction process, we keep track of these nodes and the root. Subsequently, we use the ambiguous terms and their
root names to check the accuracy of paths in the hand labeled data containing them.
As presented in \tabref{tbl:amlist}, there is about a half reduction in error for ambiguous paths using \algname{sap}.
This supports our claim about superiority of \algname{sap} on node disambiguation.

\begin{table}
\centering
\begin{tabular}{|c|l|}
\hline
\scriptsize{Approach} & \multicolumn{1}{c|}{\scriptsize{Incorrect Path}} \\
\hline
\algname{sap} & \scriptsize{anim/\textbf{other anim}/mara} \\
\hline
\algname{sap} & \scriptsize{world/landscap/\textbf{architectur}/scarborough} \\
\hline
\algname{sap} & \scriptsize{world/\textbf{scotland}/through viewfind} \\
\hline
\algname{sap} & \scriptsize{europ/\textbf{franc}/flight to} \\
\hline
\algname{sig} & \scriptsize{anim/pet/\textbf{chester}/chester zoo} \\
\hline
\algname{sig} & \scriptsize{bird/\textbf{turkei}/antalya} \\
\hline
\algname{sig} & \scriptsize{bird/\textbf{turkei}/ephesu} \\
\hline
\algname{sig} & \scriptsize{fauna/\textbf{underwat}/destin}\\
\hline
\algname{sig} & \scriptsize{south africa/\textbf{safari}/isla paulino}\\
\hline
\algname{sig} & \scriptsize{south africa/\textbf{safari}/la flore}\\
\hline
\algname{sig} & \scriptsize{sport/\textbf{golf}/adamst} \\
\hline
\algname{sig} & \scriptsize{sport/\textbf{ski}/cloud/other/new year} \\
\hline
\algname{sig} & \scriptsize{world/canada/\textbf{victoria}/melbourn} \\
\hline

\end{tabular}
\caption{The table lists all incorrect paths caused by possibly ambiguous nodes, which are in bold.}
\label{tbl:amlist}
\end{table}

In all, the proposed approach, \algname{sap}, has several advantages over the
baseline, \algname{sig}. First, it exploits both structure information and tag statistics to combine relevant saplings, 
which can produce more comprehensive folksonomies as well as resolve ambiguity of the concept
names. Second, it allows similar concepts to appear multiple times within the same
hierarchy. For example, \algname{sap}
allows the \term{anim} folksonomy to have both \term{anim}
$\rightarrow$ \term{pet} $\rightarrow$ \term{cat} and \term{anim}
$\rightarrow$ \term{mammal} $\rightarrow$ \term{cat} paths, while only
one of these paths is retained by \algname{sig}. Last, \algname{sap} can identify synonyms from structure (loops). We
learned the following synonyms from Flickr data: \{\term{anim},
\term{creatur}, \term{critter}, \term{all anim}, \term{wildlife}\} and
\{\term{insect}, \term{bug}\}.
\section{Related work}

Constructing ontological relations from text has long interested researchers, e.g., \cite{Hearst92,SnowWordnetBayes,YangACL09}. Many of these methods exploit linguistic patterns to infer if two keywords are related under a certain relationship. However, these approaches are not applicable to social metadata because it is usually \emph{ungrammatical} and much more \emph{inconsistent} than natural language text.

Several researchers have investigated various techniques to construct conceptual hierarchies from social metadata. Most of the previous work utilizes tag statistics as evidence. Mika~\cite{Mika07_OntoAreUs} uses a graph-based approach to construct a network of related tags, projected from either a user-tag or object-tag association graphs; then induces broader/narrower relations using betweenness centrality and set theory. Other works apply clustering techniques to tags, and use their co-occurrence statistics to produce conceptual hierarchies~\cite{Brooks06}. Heymann and Garcia-Molina~\cite{Heymann06} use centrality in the similarity graph of tags. The tag with the highest centrality is considered more abstract than one with a lower centrality; thus it should be merged to the hierarchy first, to guarantee that more abstract nodes are closer to the root. Schmitz~\cite{SchmitzTagging06} applied a statistical subsumption model~\cite{SandersonC99} to induce hierarchical relations among tags. Since these works are based on tag statistics, they are likely to suffer from the ``popularity vs. generality'' problem, where a tag may be used more frequently not because it is more general, but because it is more popular among users.

Our present work, \algname{sap}, is different from our earlier approach, \algname{sig}~\cite{www09folksonomies} in many aspects. First, \algname{sap} exploits more evidence, i.e., structure and tag statistics of personal hierarchies rather than individual relations' co-occurrence statistics as in \algname{sig}. Second, \algname{sap} is based on the relational clustering approach that incrementally attaches relevant saplings to the learned folksonomies, as \algname{sig} exhaustively determines the best path out of all possible paths from the root node to a leaf, which is computationally expensive when the learned folksonomies are deep. Last, \algname{sap} demonstrates many advantages as presented in \secref{sec:results}.

\comment{
Specifically, it filters out conflicting and noisy relations based on parent and child nodes' co-occurrence statistics; then, combines these relations into a larger folksonomy. Although this approach can bypass the ``popularity vs. generality'' problem, it does not address the ambiguity challenge. Particularly, it assumes that nodes with the similar names refer to the same concept. Hence, they are incorrectly integrated into the same folksonomies. Moreover, since individual relations are judged to be irrelevant only from the co-occurrence statistics, rather than from tag and structure information as in our present work does, many informative relations are unfortunately discarded. 
}
\comment{
\algname{sap} produces bushier trees because individual saplings will be judged relevant using structural information, rather than frequencies of individual relations as in \algname{sig}. Although \algname{sig} can remove many idiosyncratic relations, it also removes many of informative ones too.
}
The sapling merging approach described in this paper is an extension of collective relational clustering approach used for entity resolution \cite{EntityResoluteGetoor07}. That work proposed a method to identify and disambiguate entities, such as authors, that utilizes two types of evidence: intrisic and extrinsic features. Intrinsic features are associated with specific instances, such as author names, while extrinsic features are derived from structural evidence, e.g., co-authors in a citations database. Intuitively, two names refer to the same author if they are similar and their co-author names refer to the same set of authors. Analogously, we identify and disambiguate concept names from names and tags (intrinsic) and neighboring nodes' features (extrinsic). However, for efficiency reasons, we use the naive version of the relational clustering, where we directly use the features from neighbors as the extrinsic features, rather than cluster labels.

Handling mutual shortcuts by keeping the sapling which is more similar to the ancestor is similar in spirit to the minimum evolution assumption in \cite{YangACL09}. Specifically, a certain hierarchy should not have any sudden changes from a parent to its child concepts.
Our approach is also similar to several works on ontology alignment (e.g.~\cite{ontomatchbook07,UdreaGM07}). However, unlike those works, which merge a small number of deep, detailed and consistent concepts, we merge large number of noisy and shallow concepts, which are specified by different users.
\section{Conclusion}
This paper describes an approach which incrementally combines a large number of shallow hierarchies specified by different users into common, denser and deeper folksonomies. The approach addresses the challenges of learning folksonomies from social metadata and demonstrates several advantages over the previous work\comment{: disambiguating concepts and allowing similar concepts to appear at multiple places within the same folksonomy. Empirical results demonstrate that our approach can induce quite detailed folksonomies, which are also more consistent with taxonomies of the Open Directory Project}. Additionally, it is general enough for other domains, such as tags/bundles in \emph{Delicious} and files/folders in personal workspaces.

For the future work, in addition to automatically separating broader/narro\-wer from related-to relations, we would like to develop a systematic way to handle individual saplings whose child nodes are from different facets.  \comment{e.g., in the garden case} This will improve the quality of the learned folksonomies by not mixing concepts from different facets. We are also working on combining more sources of evidence such as geographical information for learning accurate folksonomies. Lastly, we would like to frame the approach in a fully probabilistic way (e.g.,~\cite{hmln08,psl09}), which provides a systematic way to combine heterogeneous evidence, and takes into account uncertainties on similarities between concepts and relations.

\section*{Acknowledgements} This material is based upon work supported by the National Science Foundation under Grant No. IIS-0812677.



\balancecolumns 
\end{document}